\documentclass[sigconf]{acmart}

\usepackage{multirow}
\usepackage[normalem]{ulem}
\usepackage{soul}
\usepackage{subcaption}
\usepackage{changepage}
\usepackage{balance}
\usepackage{xcolor}

\newenvironment{fullwidthquote}{
  \begin{adjustwidth}{0em}{0em}
  \itshape
}{
  \end{adjustwidth}
}

\AtBeginDocument{
  
}

\copyrightyear{2026}
\acmYear{2026}
\setcopyright{cc}
\setcctype{by}
\acmConference[MM '26]{Proceedings of the 34th ACM International Conference on Multimedia}{November 10--14, 2026}{Rio de Janeiro, Brazil.}
\acmBooktitle{Proceedings of the 34th ACM International Conference on Multimedia (MM '26), November 10--14, 2026, Rio de Janeiro, Brazil}
\acmISBN{979-8-4007-2213-4/2026/11}
\acmDOI{10.1145/3767308.3836458}
\begin{document}

\title{ProFocus: Interpreting Affective Experience in Artistic Images with Progressive Visual Focusing}

\author{Zhiyan Zhang}
\affiliation{
  \institution{University of Science and Technology of China}
  \city{Hefei}
  \state{Anhui}
  \country{China}}
\email{zzyhang02@gmail.com}

\author{Zicheng Yan}
\affiliation{
  \institution{University of Science and Technology of China}
  \city{Hefei}
  \state{Anhui}
  \country{China}}
\email{yzc2698772635@mail.ustc.edu.cn}

\author{Jianqi Chen}
\affiliation{
  \institution{University of Science and Technology of China}
  \city{Hefei}
  \state{Anhui}
  \country{China}}
\email{aphrora@mail.ustc.edu.cn}

\author{Peipei Song}
\authornote{Corresponding author.}
\affiliation{
  \institution{University of Science and Technology of China}
  \city{Hefei}
  \state{Anhui}
  \country{China}}
\email{beta.songpp@gmail.com}

\author{Shanshan Wang}
\affiliation{
  \institution{Anhui University}
  \city{Hefei}
  \state{Anhui}
  \country{China}}
\email{wang.shanshan@ahu.edu.cn}

\author{Xun Yang}
\affiliation{
  \institution{University of Science and Technology of China}
  \city{Hefei}
  \state{Anhui}
  \country{China}}
\email{xyang21@ustc.edu.cn}

\renewcommand{\shortauthors}{Zhiyan Zhang et al.}

\begin{abstract}
Interpreting the emotional responses triggered by images is central to achieving emotional intelligence. Compared with natural images, visual art is intentionally created to elicit emotional responses from its viewers through abstract concepts and visual metaphors, making affective interpretation particularly challenging. However, most existing methods rely on general-purpose visual embeddings (e.g., CLIP), failing to capture the nuanced cues underlying artistic emotion. To address this gap, we propose \textbf{ProFocus}, a novel framework that models affective experience in artistic images via progressive visual focusing. The key idea is to model visual representation learning inspired by a hierarchical cognitive theory of human aesthetic appreciation. Technically, ProFocus contains two core components: a Hierarchical Art Critic (HAC) and a Progressive Hint Fusion (PHF) module. HAC leverages multimodal large language models to generate structured linguistic priors at three cognitive levels--atmospheric style, narrative subjects, and concrete details--thereby translating artistic perception into coherent semantic guidance. Building upon these priors, PHF departs from conventional cross-modal fusion by sequentially injecting the hierarchical hints into visual features, enabling a progressive focusing process that mirrors human perception. This design allows the model to capture subtle affective cues and produce more faithful explanations. Extensive experiments on the ArtEmis v1.0 and v2.0 datasets demonstrate that ProFocus consistently outperforms state-of-the-art methods in both emotion recognition and affective explanation. Project page: https://github.com/Zhang-Zhiyan/ProFocus.
\end{abstract}

\ccsdesc[500]{Computing methodologies~Scene understanding}

\keywords{Affective reasoning, multimodal art understanding, hierarchical semantic fusion}
\maketitle

\begin{figure}[!h]
    \centering
    \includegraphics[width=0.95\columnwidth]{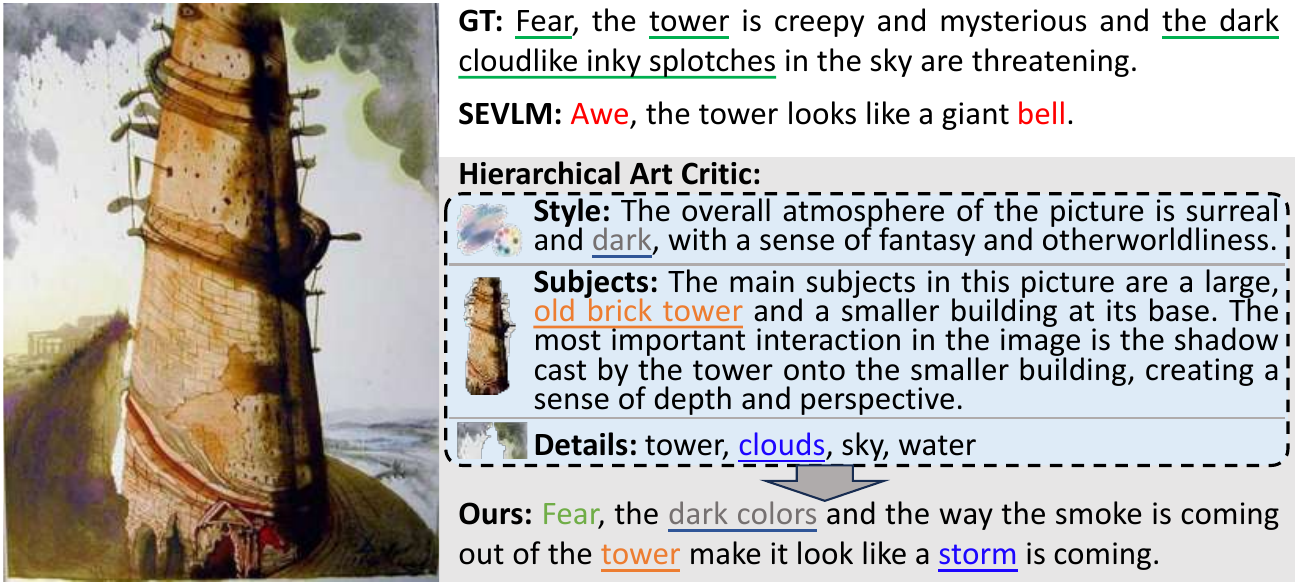}
    \vspace{-7pt}
    \caption{An example of interpreting affective experience in artistic images. The previous state-of-the-art SEVLM relies on generalist visual features. Our method enhances emotion reasoning by introducing a hierarchical art critic that analyses artwork at three explainable levels.
   }
   \vspace{-9pt}
    \label{fig:task}
\end{figure}

\section{Introduction}
\label{sec:intro}

\begin{fullwidthquote}
  ``The starting-point for all systems of aesthetics must be the personal experience of a peculiar emotion.''
    \hfill --- Clive Bell
\end{fullwidthquote}

As stated by Clive Bell in \textit{Art} \cite{bell1916art}, visual art is often created with the intent of provoking emotional reactions from its viewers. Unlike most natural images that depict objective scenes, visual art often requires viewers to go beyond content recognition and interpret how visual elements jointly contribute to an affective experience \cite{seo2004role,fernandez2021affective}. It is thus a rich medium for nuanced affective interpretation and can also inform the perceptual understanding of ordinary images \cite{artemisv1}. 
In the field of visual emotion analysis \cite{wu2025enriching,wu2025comprehensive,hu2025beyond,song2026bridging}, enabling machines to capture the emotional responses towards visual art and explain the underlying causes remains a fundamental challenge. 
This problem is essential for broader affective applications such as empathetic human-computer interaction and personalized art understanding 
\cite{hu2024psycollm,song2024emotional,picard2000affective,zhang2026affective}.

\begin{figure}[t]
    \centering
    \includegraphics[width=0.97\columnwidth]{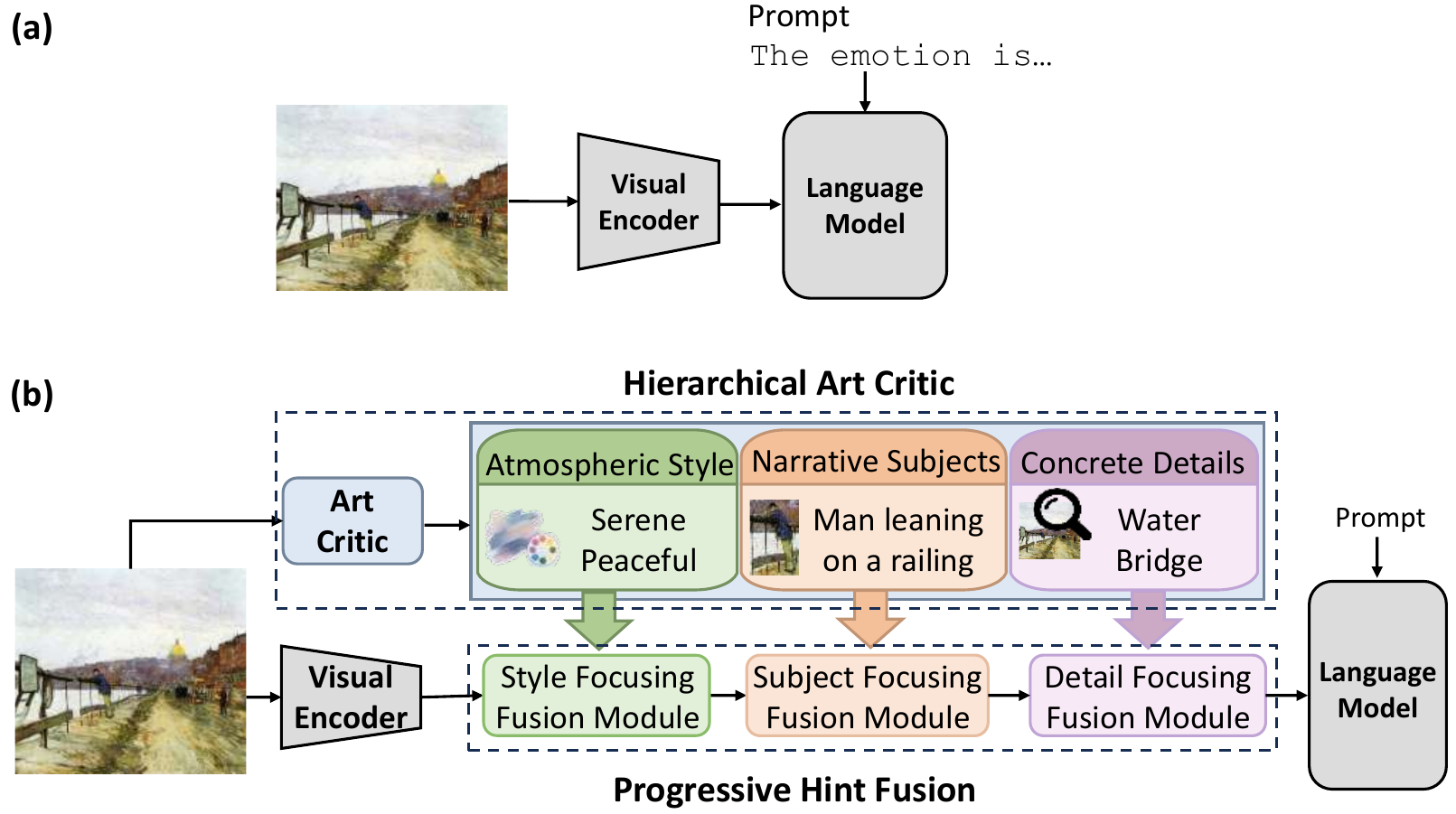}
    \vspace{-9pt}
    \caption{Pipeline comparison between (a) SEVLM and (b) the proposed ProFocus. The key innovations are the \textbf{Hierarchical Art Critic} (HAC) and \textbf{Progressive Hint Fusion} (PHF) modules.}
    \vspace{-7pt}
    \label{fig:comparison}
\end{figure}

Recent progress in Large Language Models (LLMs) \cite{bai2023qwen,achiam2023gpt,comanici2025gemini} has significantly improved multimodal emotion reasoning \cite{xie2024emovit,emotionllama,zhang2025videmo}. In the artistic domain, SEVLM \cite{sevlm} tailored a small emotional vision language model for visual art comprehension by coupling a CLIP \cite{clip} encoder with a GPT \cite{gpt2} decoder.
However, these models inherently rely on general-purpose vision encoders pre-trained on object-centric images (e.g., MS-COCO \cite{lin2014microsoft}). This leads to holistic visual representations that excel at identifying explicit objects, but are inadequate at decoupling the abstract aesthetic cues, even though the latter is pivotal for affective interpretation in visual art. 
Therefore, a profound {affective gap} exists between holistic visual features and subtle artistic interpretation.
As illustrated in Figure~\ref{fig:task}, the current SOTA model, SEVLM, is incapable of understanding the \textit{eerie} artistic style and erroneously predicts the visual emotion as ``awe''. 

To address this limitation, we draw inspiration from a cognitive theory of art appreciation \cite{leder2004model, chatterjee2014neuroaesthetics}. This theory suggests that humans' emotional response to art emerges through a \textbf{hierarchical cognitive integration}: viewers initially sense the overall style, then parse the narrative interactions of subjects, and finally dwell on concrete symbolic details. Such a progressive reasoning process is difficult to capture with holistic visual features, such as the CLIP features given by SEVLM. 
To understand the rich emotions triggered by visual art, emulating the human cognitive pathway in visual processing is a promising direction.

Based on the above insight, we propose a novel \textbf{ProFocus} for interpreting affective experiences in artistic images through progressive visual focusing. As shown in Figure \ref{fig:comparison}, ProFocus improves existing baselines (e.g., SEVLM) by explicitly aligning visual representation learning with the hierarchical process of human aesthetic perception. The core motivation is that general-purpose vision encoders are incapable of disentangling heterogeneous artistic cues, in which abstract aesthetics, compositional structure, and fine-grained symbolic details are entangled within a compact visual representation. To address this limitation, we introduce a \textbf{Hierarchical Art Critic} (HAC) that decomposes artistic perception into structured and interpretable components. It leverages a multimodal large language model to translate visual content into three levels of linguistic priors: (1) \textit{Atmospheric Style}, capturing overall style, color palette, and artistic technique; (2) \textit{Narrative Subjects}, describing main entities and their relationships; and (3) \textit{Concrete Details}, identifying specific symbolic elements.
Building upon these hierarchical priors, a remaining challenge is how to integrate them in a way that is consistent with human aesthetic appreciation. Considering that the intuitional atmosphere often establishes an emotional context for subsequent interpretation of subjects and details, we propose a \textbf{Progressive Hint Fusion} (PHF) module to perform sequential integration. It progressively injects the hierarchical priors into visual features, thereby enabling the model to preserve the dominance of holistic affect while refining attention toward semantically meaningful regions.
As shown in Figure~\ref{fig:task}, guided by the progressively focused cues (``dark'' style, ``tower'' subject, and ``smoke'' detail), the proposed ProFocus successfully predicts the emotion ``fear'' along with a visually grounded explanation.

The main contributions of this paper are summarized as follows:
\begin{itemize}
\setlength{\itemsep}{0pt}
\setlength{\parsep}{0pt}
\setlength{\parskip}{0pt}
\item Instead of using generalist visual features, we introduce a novel Progressive Visual Focusing (ProFocus) method for interpreting affective experiences in artistic images. It explicitly aligns visual processing with the human cognitive pathway.
\item We design a Hierarchical Art Critic (HAC) and a Progressive Hint Fusion (PHF) module to tackle artistic affective interpreting, which decompose the entangled visual aesthetics into hierarchical cognitive layers and incorporate those hints into the model progressively.
\item Extensive experiments on the ArtEmis v1.0 and ArtEmis v2.0 datasets validate that our proposed ProFocus consistently outperforms strong baselines and competitive LLM-based models. It improves both emotion prediction and affective explanations.
\end{itemize}

\begin{figure*}[t]
    \centering
    \includegraphics[width=0.99\textwidth]{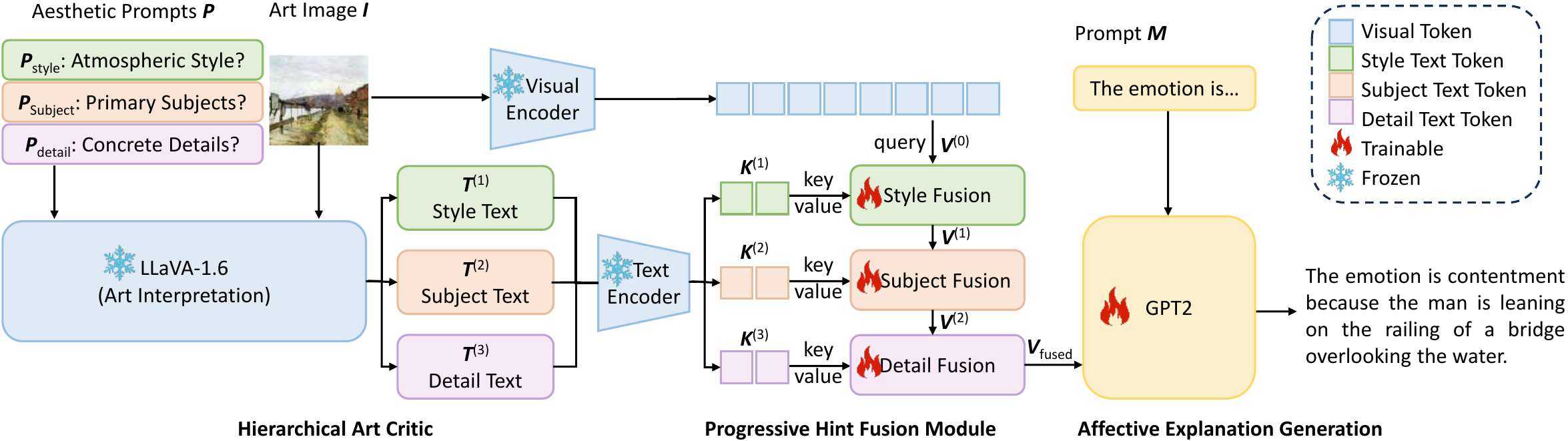}
    \vspace{-7pt}
    \caption{Overview of the proposed ProFocus framework. (1) The Hierarchical Art Critic explicitly decouples abstract visual aesthetics into structured linguistic priors across three cognitive levels. (2) The Progressive Hint Fusion module injects these priors into visual features via semantic-guided visual focusing in a coarse-to-fine sequence. (3) The hierarchically refined features are fed into a language decoder to bridge the affective gap and generate subjective affective explanations.}
    \vspace{-7pt}
    \label{fig:framework}
\end{figure*}

\section{Related Work}
\label{sec:related_work}

\subsection{Visual Art Comprehension}
The intersection of computer vision and art history \cite{gooch2002artistic,cai2015cross,wu2025review} has attracted increasing attention. Early computational aesthetics primarily focused on objective classification, utilizing hand-crafted features or Convolutional Neural Networks (CNNs) \cite{o2015introduction} to categorize artworks by artist, style, or period \cite{gatys2015neural, mazzone2019art}. To move beyond low-level attributes, the SemArt dataset \cite{garcia2018read} aligned fine-art paintings with artistic comments, enabling multi-modal retrieval and semantic art understanding.

Recently, the focus has shifted toward subjective affective interpretation. The ArtEmis datasets \cite{artemisv1,artemisv2} marked a milestone by providing large-scale emotion annotations and natural language explanations for artworks. However, extracting affective features remains challenging. Unlike natural photographs \cite{Affection}, artworks communicate through implicit metaphors, non-literal colors, and abstract brushwork \cite{barr2019cubism}. Existing models often treat paintings as generic images, applying general-purpose vision encoders that fail to disentangle the nuanced, hierarchical aesthetic cues essential for profound art comprehension.

\subsection{Visual Emotion Analysis}
Visual Emotion Analysis (VEA) traditionally formulates emotion prediction \cite{yang2023emoset,xu2022mdan} as mapping visual stimuli to discrete categories \cite{ekman} or continuous dimensional spaces \cite{mohammad2018obtaining}. To bridge the ``affective gap'' between pixels and emotions, researchers employ various feature extractors, ranging from CNN embeddings \cite{yang2018weakly, yang2021stimuli} to attention-based models targeting emotion-specific regions like facial expressions \cite{danvevcek2022emoca, krumhuber2023role}.

While effective on natural images, these models face two major limitations in the artistic domain. First, they lack interpretability, outputting static emotion labels without rationales. Second, they rely on implicitly entangled visual features. In art, emotion often reflects a holistic integration of atmospheric tones and symbolic details rather than isolated local objects. Simple attention over raw visual tokens may struggle to capture such complex aesthetic metaphors, motivating explicit semantic grounding.

\subsection{Explainable Affective Reasoning}
To address traditional VEA's lack of interpretability, recent research shifted towards explainable affective reasoning. While early works \cite{artemisv1} treated emotion classification and rationale generation disjointly, end-to-end frameworks like SEVLM \cite{sevlm} directly integrated frozen visual encoders with language decoders. Recently, the rapid emergence of Multimodal Large Language Models (MLLMs) \cite{zeng2025glm,LLaVA-OneVision,chen2024internvl} popularized generative paradigms for complex emotion understanding \cite{han2025benchmarking, song2023emotion, zhang2026stimuli, zhang2026benchmarking}. Advanced models like EmoVIT \cite{xie2024emovit} and VidEmo \cite{zhang2025videmo} utilize massive vision-language architectures, frequently employing Chain-of-Thought (CoT) prompting \cite{wei2022chain,shao2024visual,zhang2026cmmcot,han2026omni} to articulate intermediate reasoning steps before reaching emotional conclusions. 

Despite succeeding on natural images and videos, applying general-purpose MLLMs directly to visual art remains sub-optimal. Artworks often lack explicit facial expressions or dynamic actions; their emotional impact relies heavily on the complex interplay of abstract style, composition, and symbolic metaphors. Generic CoT prompting often fails to disentangle these nuanced aesthetic cues. To bridge this affective gap, our work introduces an aesthetically-grounded cognitive pathway, explicitly decoupling the artistic visual space into hierarchical semantic priors before conducting affective reasoning.

\section{Methodology}
\label{sec:methodology}

The overall architecture of the proposed framework is shown in Figure~\ref{fig:framework}. Given an artwork image $I$, our goal is to predict an emotion label $E$ and generate a corresponding explanation $X$. To bridge the semantic gap in abstract art understanding, we propose a framework with three components: (1) a Hierarchical Art Critic that uses LLaVA-1.6 to emulate human aesthetic cognition; (2) a  Progressive Hint Fusion module that progressively injects these hierarchical priors into visual features; and (3) an Affective Explanation Generation module that predicts emotions and generates rationales from the fused representations.

\subsection{Baseline Architecture}
Our method builds on the SEVLM \cite{sevlm}, a baseline designed for artistic emotions. It contains an image encoder, a text encoder, and a language decoder. Formally, SEVLM uses a frozen CLIP \cite{clip} as the image encoder to extract visual features $\mathbf{f}^I \in \mathbb{R}^{K \times d_v}$ from the raw image $I$, where $K$ is the number of spatial patches. These $K$ patches preserve localized structural information but do not provide explicit affective semantics. The textual input consists of an emotion prompt $M$ (``\texttt{The emotion is \_}'') and an explanation $X$. These inputs are converted into embeddings by summing word embeddings, position embeddings, and segment embeddings:
\begin{equation}
    \mathbf{f}^S = \text{Embed}(M) \oplus \text{Embed}(X) \in \mathbb{R}^{L \times d_s},
\end{equation}
where $\oplus$ denotes concatenation along the sequence dimension.
Finally, a GPT-based language decoder generates the output sequence. It takes the concatenated visual and textual features as input. In the original SEVLM, the hidden states are computed through cross-attention:
\begin{equation}
    \mathbf{H}_{emo}, \mathbf{H}_{exp} = \text{GPT2Decoder}(\mathbf{f}^S, \mathbf{f}^I),
\end{equation}
where $\mathbf{H}_{emo}$ and $\mathbf{H}_{exp}$ denote the hidden states for emotion classification and explanation generation, respectively.

To retain interpretable emotion analysis, SEVLM is optimized end-to-end with a multi-task objective:
\begin{equation}
    \mathcal{L}_{total} = \mathcal{L}_{lm} + \lambda_1 \mathcal{L}_{vad} + \lambda_2 \mathcal{L}_{cl},
\end{equation}
where $\mathcal{L}_{lm}$ is the standard cross-entropy language modeling loss. $\mathcal{L}_{vad}$ is a VAD-consistency loss that computes the Valence-Arousal-Dominance vectors of the generated words and aligns them with the ground-truth VAD \cite{mohammad2018obtaining} vectors through Mean Squared Error (MSE), thereby preserving appropriate emotional intensity. $\mathcal{L}_{cl}$ is a contrastive alignment loss that pulls the global representations of the input image, the emotion category, and the explanation closer in a shared latent space.

\textbf{Limitation.} A key limitation of this baseline is that the visual feature $\mathbf{f}^I$ is primarily extracted by general-purpose encoders optimized for object-centric natural images. In such holistic representations, heterogeneous artistic cues—such as abstract aesthetics, narrative structures, and concrete symbolic details—are tightly entangled within a single feature space. This hinders the model from decoding the nuanced, multi-layered visual metaphors essential for affective reasoning. To bridge this affective gap, our proposed framework reduces the reliance on implicitly entangled embeddings. Instead, we aim to reconstruct the visual reasoning pathway by explicitly aligning representation learning with the human cognitive process of art appreciation.

\subsection{Hierarchical Art Critic}
\label{sec:interpretation}
To explicitly decouple the entangled visual cues, we introduce a Hierarchical Art Critic (HAC). The core motivation is that raw visual features encode artistic semantics only implicitly. Although generic vision encoders preserve spatial layouts, they struggle to disentangle atmospheric, narrative, and symbolic cues in the artwork. As a result, the language decoder often struggles to articulate \textit{why} a specific emotion is elicited. 
To bridge this gap, HAC emulates the hierarchical nature of human art appreciation \cite{leder2004model}.
Rather than relying on unstructured visual tokens, HAC leverages a multimodal large language model (LLaVA-1.6 \cite{liu2024llavanext}) as a virtual art critic. It emulates the aforementioned coarse-to-fine perception process, translating latent visual aesthetics into explicit, multi-level linguistic priors before feature fusion.

\textbf{Hierarchical Decomposition.}
As shown in Figure~\ref{fig:framework}, we decompose the artistic perception into three progressive cognitive layers. We define a prompt set $\mathcal{P} = \{P_{style}, P_{subj}, P_{detail}\}$ to query the MLLM. First, we request an \textit{Atmospheric Style} description to capture the initial impression, focusing exclusively on the overall artistic style, color palette, and painting techniques. This level provides the global semantic context and serves as the foundational aesthetic prior for the subsequent visual representation. Second, we shift to \textit{Narrative Subjects} to identify the main figures, objects, and their spatial or semantic relations, thereby organizing the artwork into a more interpretable narrative structure. This level bridges the global atmosphere and local evidence by specifying \emph{what} entities are present and \emph{how} they interact. Finally, we query \textit{Concrete Details} to extract fine-grained visual evidence, such as background elements, symbolic anchors, and localized attributes that directly support the emotional interpretation. In this way, the three layers form a progressive semantic chain: from holistic aesthetic impression, to structured narrative understanding, and down to concrete visual evidence.

Formally, the generation of hierarchical textual descriptions $\mathcal{T}$ is written as:
\begin{equation}
    \mathcal{T} = \{T^{(1)}, T^{(2)}, T^{(3)}\} = \text{LLaVA}(I, \mathcal{P}),
\end{equation}
where $T^{(1)}$, $T^{(2)}$, and $T^{(3)}$ correspond to the style, subject, and detail descriptions, respectively. By explicitly aligning with the aforementioned cognitive pathway, this hierarchical formulation provides a structured and task-relevant decomposition of the artistic content, laying a precise semantic foundation for the subsequent feature fusion.

Furthermore, this strategy significantly enhances the interpretability of the overall framework. Each textual description acts as a distinct semantic anchor: $T^{(1)}$ establishes the overarching aesthetic context, $T^{(2)}$ clarifies the narrative organization, and $T^{(3)}$ pinpoints fine-grained visual evidence. Consequently, these intermediate outputs are not only crucial for regularizing downstream visual representations, but they also provide an explicit semantic trace of how the model deconstructs an artwork before generating the final affective explanation.

\textbf{Text Encoding.} To utilize these descriptions in the subsequent fusion module, a pre-trained text encoder maps each description $T^{(l)}$ into a sequence of embeddings:
\begin{equation}
    \mathbf{K}^{(l)} = \text{TextEncoder}(T^{(l)}) \in \mathbb{R}^{L_t \times d}, \quad l \in \{1, 2, 3\}.
\end{equation}
These embeddings $\mathbf{K}^{(l)}$ act as explicit linguistic priors. By converting abstract aesthetic elements into explicit linguistic tokens, they provide a structured vocabulary that the visual features can subsequently attend to. More importantly, the three groups of textual embeddings encapsulate distinct hierarchical hints: $\mathbf{K}^{(1)}$ emphasizes the atmospheric style, $\mathbf{K}^{(2)}$ captures narrative-level subjects, and $\mathbf{K}^{(3)}$ focuses on concrete visual evidence. This hierarchical separation is essential for the subsequent fusion stage, where different levels of semantic priors are progressively aligned with localized image regions in a coarse-to-fine manner. In this sense, the HAC serves as the semantic front end of our framework, transforming unstructured artistic perception into a layered and fusion-ready representation space.

\begin{figure}[t]
    \centering
    \includegraphics[width=0.7\columnwidth]{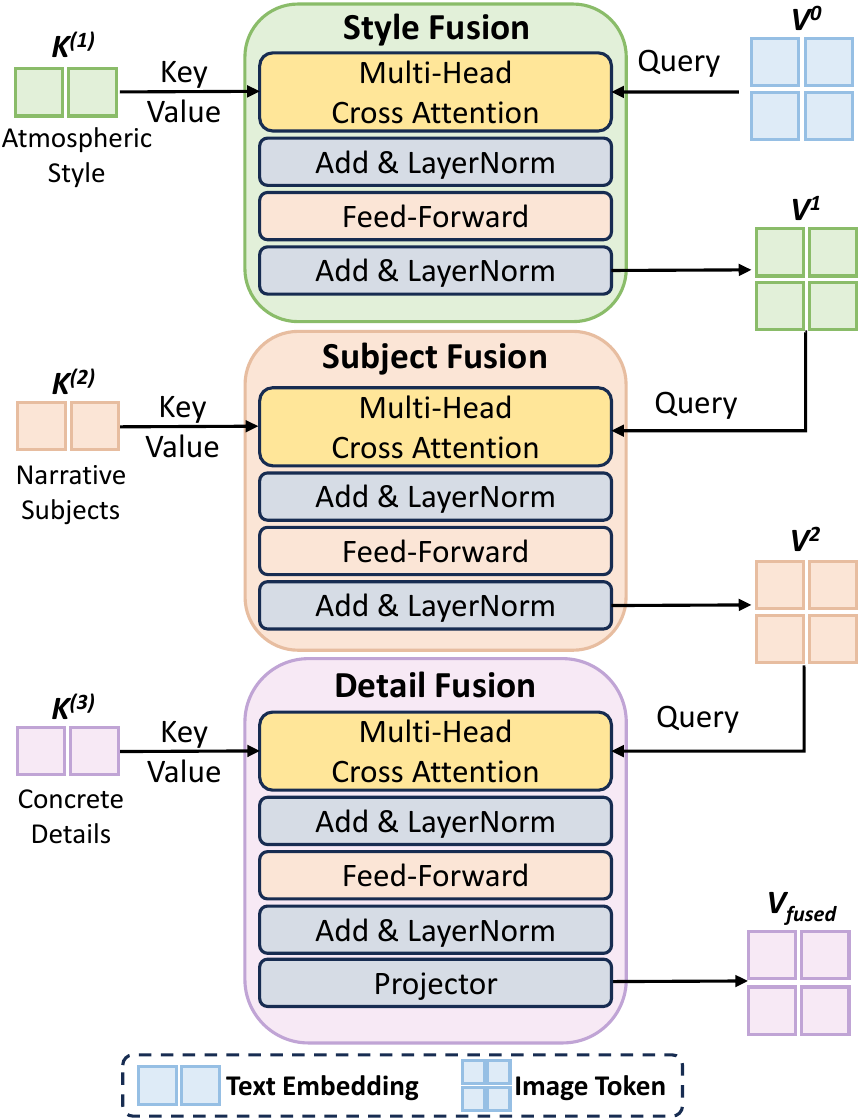}
    \vspace{-7pt}
    \caption{Detailed architecture of the Progressive Hint Fusion. Visual tokens act as queries to retrieve information from hierarchical textual keys and values across three stages.}
    \vspace{-7pt}
    \label{fig:fusion}
\end{figure}

\subsection{Progressive Hint Fusion}
\label{sec:fusion}
Building upon the hierarchical linguistic priors extracted by the HAC, a subsequent challenge lies in integrating them into the visual representation in a manner consistent with human aesthetic perception. To achieve this, we design the \textbf{Progressive Hint Fusion (PHF)} module, as illustrated in Figure~\ref{fig:fusion}. Rather than a one-step fusion, its core mechanism is a sequential spatial-to-semantic focusing process that maps localized visual patches to abstract linguistic concepts in a coarse-to-fine order.

\textbf{Fusion Block.}
The core function $\Phi_l(\cdot)$ of the fusion block is implemented with Multi-Head Cross-Attention (MCA). At the $l$-th stage, the visual features $\mathbf{V}^{(l-1)} \in \mathbb{R}^{K \times D}$ act as the \textbf{Queries} ($\mathbf{Q}$), representing $K$ spatial image patches. The textual priors $\mathbf{K}^{(l)} \in \mathbb{R}^{L_t \times D}$ act as both \textbf{Keys} ($\mathbf{K}$) and \textbf{Values} ($\mathbf{V}$), representing $L_t$ explicit semantic concepts. Formally, the attention operation is defined as:
\begin{equation}
    \mathbf{Q} = \mathbf{V}^{(l-1)}\mathbf{W}_Q, \quad \mathbf{K} = \mathbf{K}^{(l)}\mathbf{W}_K, \quad \mathbf{V} = \mathbf{K}^{(l)}\mathbf{W}_V
\end{equation}
\begin{equation}
    \text{MCA}(\mathbf{Q}, \mathbf{K}, \mathbf{V}) = \text{Softmax}\left(\frac{\mathbf{Q}\mathbf{K}^\top}{\sqrt{d_k}}\right)\mathbf{V},
\end{equation}
where $\mathbf{W}_Q$, $\mathbf{W}_K$, and $\mathbf{W}_V$ are learnable projection matrices.
Instead of naive feature concatenation, the MCA output $\text{MCA}(\mathbf{Q},\mathbf{K},\mathbf{V})\in \mathbb{R}^{K\times D}$ acts as a semantic-guided visual focusing mechanism. Its attention weights capture the affinity between each spatial patch and textual token. Driven by the design, this mechanism progressively mirrors the human cognitive pathway of aesthetic appreciation. Specifically, the initial stage establishes a global aesthetic context. Subsequently, when fusing narrative subjects and concrete details, these attention weights refine the visual focus toward specific semantically meaningful regions (e.g., a patch with chaotic strokes explicitly attending to the token ``stormy''). This ensures that the model preserves the dominance of the holistic affect while capturing fine-grained artistic evidence.

\begin{table*}[t]
\centering
\caption{
Performance comparison with state-of-the-art methods on the ArtEmis v1.0 and v2.0 datasets.
The best results are highlighted in \textbf{bold}.
The first three methods are emotion-focused multimodal large language models adapted to the image-based setting, whereas the remaining four methods are task-specific models trained on the ArtEmis datasets.
For Emotion-Llama, the backbone is denoted as \textit{Hybrid-LLaMA2}, where \textit{Hybrid} refers to a multi-view multimodal encoder composed of HuBERT, a MAE-based local ViT encoder, VideoMAE, and EVA, and LLaMA2 serves as the decoder.
}
\vspace{-7pt}
\label{tab:main_results}
\resizebox{0.98\textwidth}{!}{
\begin{tabular}{l|c|ccccccc|ccccccc}
\toprule
\multirow{2}{*}{\textbf{Method}} & \multirow{2}{*}{\textbf{Backbone}} & \multicolumn{7}{c|}{\textbf{ArtEmis v1.0}} & \multicolumn{7}{c}{\textbf{ArtEmis v2.0}} \\
 &  & \textbf{ACC} & \textbf{B@1} & \textbf{B@2} & \textbf{B@3} & \textbf{B@4} & \textbf{M} & \textbf{R} & \textbf{ACC} & \textbf{B@1} & \textbf{B@2} & \textbf{B@3} & \textbf{B@4} & \textbf{M} & \textbf{R} \\ \midrule
Emotion-Llama \cite{emotionllama} & Hybrid-LLaMA2 & 33.4 & 20.4 & 9.8 & 2.7 & 1.4 & 20.1 & 15.6 & 16.3 & 14.2 & 7.4 & 3.5 & 1.7 & 19.3 & 15.6 \\
AffectGPT \cite{lian2025affectgpt} & CLIP-Qwen2.5 & 35.0 & 33.5 & 11.6 & 2.8 & 1.6 & 22.4 & 16.0 & 15.4 & 27.6 & 9.6 & 3.6 & 1.8 & 20.5 & 13.8 \\
VidEmo \cite{zhang2025videmo} & Qwen2.5-VL & 45.9 & 34.4 & 13.0 & 4.9 & 2.2 & 20.2 & 19.5 & 23.2 & 28.5 & 10.9 & 4.3 & 2.1 & 18.0 & 17.4 \\ \midrule
M2 \cite{artemisv1} & Trans-Trans & 60.2 & 51.1 & 28.2 & 15.4 & 9.0 & 13.7 & 28.6 & - & - & - & - & - & - & - \\
SAT \cite{artemisv1} & CNN-LSTM & 60.2 & 52.0 & 28.0 & 14.6 & 7.9 & 13.4 & 29.4 & 43.3 & 48.7 & 25.3 & 13.2 & 7.3 & 12.8 & 27.2 \\
NLX-GPT2 \cite{nlxgpt} & CLIP-GPT2 & 63.6 & 53.8 & 29.9 & 16.3 & 9.3 & 13.8 & 30.3 & 41.8 & 51.7 & 30.6 & 17.6 & 10.5 & 13.8 & 30.7 \\
SEVLM \cite{sevlm} & CLIP-GPT2 & 63.9 & 54.0 & 29.8 & 16.0 & 8.9 & 13.7 & 30.2 & 41.0 & 51.8 & 30.5 & 17.3 & 10.0 & 13.7 & 30.6 \\ \midrule
\textbf{ProFocus} & CLIP-GPT2 & \textbf{66.1} & \textbf{55.3} & \textbf{31.2} & \textbf{17.1} & \textbf{9.8} & \textbf{14.2} & \textbf{30.8} & \textbf{43.8} & \textbf{52.4} & \textbf{31.2} & \textbf{18.0} & \textbf{10.8} & \textbf{14.1} & \textbf{30.8} \\ \bottomrule
\end{tabular}
}
\vspace{0pt}
\end{table*}

\textbf{Progressive Injection.}
Standard approaches often fuse multimodal features simultaneously \cite{cheng2026mojitomodaljointlearning}, which can easily cause salient visual objects to overshadow the artwork's subtle, overarching mood. To overcome this, we stack three fusion blocks enforcing a strict coarse-to-fine order. Let $\mathbf{V}^{(0)} = \mathbf{f}^I$ denote the initial visual features. The hierarchical fusion proceeds recursively:
\begin{equation}
    \mathbf{V}^{(l)} = \Phi_l(\mathbf{V}^{(l-1)}, \mathbf{K}^{(l)}), \quad \text{for } l = 1, 2, 3.
\end{equation}
To ensure training stability and prevent visual features from being overwhelmed by textual noise, we adopt a \textbf{Post-LayerNorm} architecture with residual connections:
\begin{align}
    \mathbf{Z} &= \text{LN}\big(\mathbf{V}^{(l-1)} + \text{Dropout}(\text{MCA}(\mathbf{Q}, \mathbf{K}, \mathbf{V}))\big), \label{eq:mca_res}\\
    \mathbf{V}^{(l)} &= \text{LN}\big(\mathbf{Z} + \text{Dropout}(\text{FFN}(\mathbf{Z}))\big). \label{eq:ffn_res}
\end{align}
The residual connection in Eq.~\ref{eq:mca_res} is crucial, as it preserves the original spatial inductive biases of the image to serve as structural anchors while progressively absorbing semantic knowledge. The Feed-Forward Network (FFN) in Eq.~\ref{eq:ffn_res} further increases representational capacity to blend the heterogeneous multimodal information.

Ultimately, $\mathbf{V}^{(1)}$ absorbs the atmospheric style, $\mathbf{V}^{(2)}$ incorporates narrative subjects, and $\mathbf{V}^{(3)}$ is refined with concrete details. To map the hierarchically fused features into a decoder-friendly latent space, the output of the final stage is passed through a projector and layer normalization:
\begin{equation}
    \mathbf{V}_{fused} = \text{LN}(\text{Projector}(\mathbf{V}^{(3)})).
\end{equation}
This $\mathbf{V}_{fused}$ serves as a highly discriminative, aesthetically-grounded input for the language decoder.

\subsection{Affective Explanation Generation}
The final stage performs \textbf{Affective Explanation Generation}, which translates the decoupled aesthetic perception into subjective emotional reasoning. Instead of relying on implicitly entangled raw features $\mathbf{f}^I$, the GPT language decoder receives the hierarchically refined features $\mathbf{V}_{fused}$. 

The GPT language decoder takes the textual embeddings $\mathbf{f}^S$ together with $\mathbf{V}_{fused}$ as input. It operates autoregressively to first predict the emotion label $\hat{E}$ and then generate the explanation $\hat{X}$:
\begin{equation}
    \hat{E}, \hat{X} = \text{GPT2Decoder}(\mathbf{f}^S, \mathbf{V}_{fused}).
\end{equation}

By conditioning on $\mathbf{V}_{fused}$, the decoder can exploit the explicitly injected semantic cues, namely style, subjects, and details, to generate more rational and visually grounded explanations. The entire framework is trained end-to-end with the same multi-task objective $\mathcal{L}_{total}$ introduced earlier. Joint optimization over $\mathcal{L}_{lm}$, $\mathcal{L}_{vad}$, and $\mathcal{L}_{cl}$ encourages the Progressive Hint Fusion module to extract and route the most relevant linguistic priors, thereby improving both emotion prediction accuracy and visual-semantic alignment.

\begin{figure*}[t]
    \centering
    \includegraphics[width=0.98\textwidth]{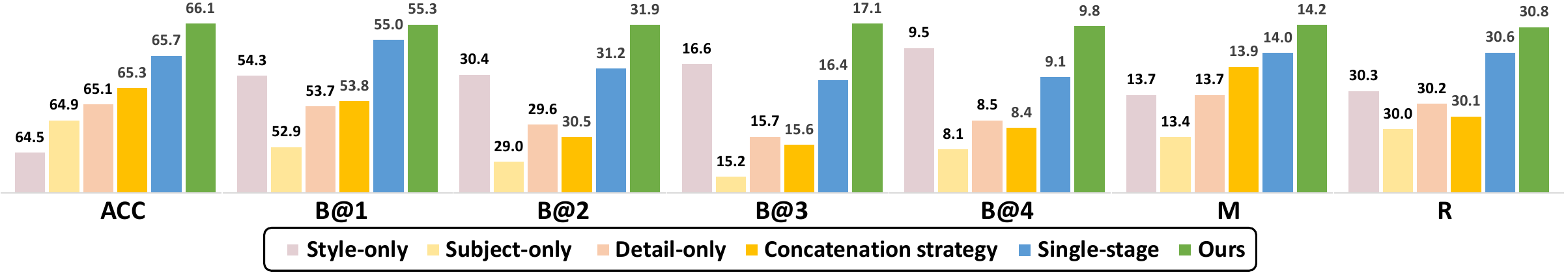}
    \vspace{-7pt}
    \caption{Ablation study on the ArtEmis v1.0 test set. We evaluate three single-prior variants (\textit{Style-only}, \textit{Subject-only}, and \textit{Detail-only}), a \textit{Concatenation} strategy that replaces the proposed attention-based fusion, and a \textit{Single-stage} variant that removes the hierarchical multi-stage design. The full model consistently performs best on ACC, BLEU@1--4, METEOR, and ROUGE-L, which confirms the value of complementary semantic priors and progressive coarse-to-fine fusion.}
    \label{fig:ablation}
    \vspace{-5pt}
\end{figure*}

\section{Experiments}
\label{sec:experiments}

\subsection{Experimental Settings}

\noindent\textbf{Datasets.} We conduct experiments on two benchmark datasets for affective reasoning. \textbf{ArtEmis v1.0} \cite{artemisv1} contains 454,684 affective explanations for 80,031 artworks from WikiArt. \textbf{ArtEmis v2.0} \cite{artemisv2} is an upgraded version of ArtEmis v1.0 in which 52,933 artworks were re-annotated to reduce emotional bias, adding 260,533 additional annotations to the dataset. Both datasets categorize emotions into nine classes based on Ekman's theory \cite{ekman}: \textit{anger}, \textit{disgust}, \textit{fear}, and \textit{sadness} as negative emotions; \textit{amusement}, \textit{awe}, \textit{contentment}, and \textit{excitement} as positive emotions; and a \textit{something else} category. Following prior work \cite{artemisv1,artemisv2,sevlm}, we adopt the standard split of 85\% for training, 5\% for validation, and 10\% for testing.

\noindent\textbf{Evaluation Metrics.} For emotion recognition, we report \textbf{Accuracy (ACC)}, which measures the consistency between predicted and ground-truth dominant emotions. For explanation generation, we use standard captioning metrics, including \textbf{BLEU@$n$} (B@1--4), \textbf{METEOR (M)}, and \textbf{ROUGE-L (R)}, to evaluate semantic relevance and linguistic fluency.

\noindent\textbf{Implementation Details.} The proposed framework is implemented in PyTorch. The visual backbone uses a pre-trained CLIP-ViT-B/16 \cite{clip} encoder and remains frozen during training. The Affective Explanation Generation module is based on GPT2 \cite{gpt2}. For the Hierarchical Art Critic, we use LLaVA-1.6-7B \cite{liu2024llavanext} to generate hierarchical textual descriptions offline, and we encode them into 768-dimensional embeddings. The Progressive Hint Fusion Module contains three stages, a hidden size of 768, 12 attention heads, and a dropout rate of 0.05. We train the model with the AdamW \cite{adamw} optimizer using a learning rate of $2e{-5}$ and a batch size of 8. Training lasts 20 epochs and requires approximately 16 hours on two NVIDIA Tesla V100 GPUs with 32 GB memory each.

\subsection{Experimental Results}

We compare the proposed method with existing state-of-the-art models, including M2 \cite{artemisv1}, SAT \cite{artemisv1}, NLX-GPT2 \cite{nlxgpt}, and the SEVLM baseline \cite{sevlm}. In addition, we evaluate three recent video emotion understanding models based on LLMs: Emotion-Llama \cite{emotionllama}, AffectGPT \cite{lian2025affectgpt}, and VidEmo \cite{zhang2025videmo}. For fair evaluation in the image-based affective reasoning setting, we adapt these video models by providing a single static frame as input. Table~\ref{tab:main_results} reports the quantitative results on the ArtEmis v1.0 and v2.0 test sets.

In terms of \textbf{Emotion Recognition}, the proposed method establishes a new state-of-the-art on both datasets. On ArtEmis v1.0, the framework reaches an accuracy of \textbf{66.1\%}, which improves on the SEVLM baseline by 2.2 percentage points. A similar trend appears on ArtEmis v2.0, where the proposed method achieves \textbf{43.8\%} accuracy compared with 41.0\% for the baseline. By contrast, the recent LLM-based video emotion models perform poorly on this task. For example, VidEmo and AffectGPT achieve only 45.9\% and 35.0\% accuracy on ArtEmis v1.0, respectively. This gap suggests that temporal modeling or general-purpose explicit cues, such as facial expressions or actions, are insufficient for understanding static abstract art. These consistent gains indicate that progressive injection of semantic priors, from global style to fine-grained details, substantially improves the ability of the model to distinguish subtle and implicit emotional cues in artworks.

Regarding \textbf{Emotion Explanation}, the proposed method also produces more fluent, better aligned, and more visually grounded rationales. On ArtEmis v1.0, the model achieves a BLEU@4 score of \textbf{9.8} and a ROUGE-L score of \textbf{30.8}, clearly surpassing the SEVLM baseline. The LLM-based video emotion models, such as AffectGPT, show unusually high METEOR scores, for example 22.4 on v1.0, likely because they generate long and weakly constrained outputs with broad vocabulary coverage. However, these models perform poorly on exact semantic alignment and structural fluency, as reflected by sharp degradation in BLEU and ROUGE-L scores, such as a BLEU@4 score of only 1.6. On the more challenging ArtEmis v2.0 split, the proposed method continues to lead, achieving a BLEU@4 score of 10.8 and a ROUGE-L score of 30.8. These results indicate that, compared with unconstrained generation from large LLMs, the proposed hierarchical semantic fusion effectively bridges the affective gap, enabling the decoder to produce concise, convincing, and well-grounded affective explanations.

\subsection{Ablation Study}
\label{sec:ablation}

To examine the contribution of each component in the proposed framework, we conduct extensive ablation studies on the ArtEmis v1.0 dataset. As illustrated in Figure~\ref{fig:ablation}, we focus on the impact of distinct semantic priors, the effectiveness of the attention-based fusion mechanism, and the necessity of the multi-stage architecture.

\noindent\textbf{Impact of Progressive Hint Fusion Module.}
We first examine whether different semantic priors are complementary by forcing the fusion module to use only one type of textual guidance, repeated across all three stages. As shown in Figure~\ref{fig:ablation}, relying on a single prior limits overall performance. The \textit{detail}-only variant achieves an accuracy of 65.1\%, which is higher than the 64.5\% of the \textit{style}-only variant, suggesting that fine-grained visual evidence is especially informative for emotion classification. However, for explanation generation, the \textit{style}-only variant reaches a BLEU@4 score of 9.5, whereas the \textit{detail}-only variant reaches only 8.5. This contrast indicates that different semantic priors play complementary roles: concrete details help emotion recognition, whereas atmospheric style is especially important for coherent and high-quality explanations. Integrating these complementary views yields the best overall performance.

\noindent\textbf{Effectiveness of Attention-Based Fusion.}
To validate the effectiveness of the Progressive Hint Fusion, we replace the cross-attention-based design with a simple concatenation strategy. In this variant, the three textual embeddings are concatenated directly with visual features and then fed to the decoder through a linear projection. This concatenation strategy yields an ACC of 65.3\% and a BLEU@4 score of 8.4. In contrast, the proposed attention-based fusion improves the ACC to 66.1\% and the BLEU@4 score to 9.8. The lower scores suggest that naive concatenation is less effective for semantic integration. By contrast, the semantic-guided visual focusing mechanism allows the model to selectively absorb the most relevant linguistic priors, resulting in substantially better explanation generation.

\begin{table}[t]
\centering
\caption{Ablation study on the order of semantic text injection in the three-stage fusion module. The proposed coarse-to-fine order yields the best results.}
\vspace{-7pt}
\label{tab:ablation_order}
\resizebox{\columnwidth}{!}{
\begin{tabular}{l|ccccccc}
\toprule
\textbf{Fusion Order} & \textbf{ACC} & \textbf{B@1} & \textbf{B@2} & \textbf{B@3} & \textbf{B@4} & \textbf{M} & \textbf{R} \\ \midrule
Style $\rightarrow$ Detail $\rightarrow$ Subject & 65.7 & 53.9 & 30.1 & 16.3 & 9.1 & 13.6 & 30.3 \\
Subject $\rightarrow$ Style $\rightarrow$ Detail & 64.9 & 54.8 & 30.5 & 16.5 & 9.3 & 14.0 & 30.4 \\
Detail $\rightarrow$ Style $\rightarrow$ Subject & 66.0 & 54.5 & 30.3 & 16.2 & 9.0 & 14.0 & 30.4 \\
Subject $\rightarrow$ Detail $\rightarrow$ Style & 64.9 & 54.3 & 30.1 & 16.1 & 9.0 & 13.7 & 30.5 \\
Detail $\rightarrow$ Subject $\rightarrow$ Style & 64.5 & 54.2 & 30.0 & 16.3 & 9.2 & 13.4 & 30.5 \\ \midrule
\textbf{Style $\rightarrow$ Subject $\rightarrow$ Detail} & \textbf{66.1} & \textbf{55.3} & \textbf{31.2} & \textbf{17.1} & \textbf{9.8} & \textbf{14.2} & \textbf{30.8} \\ \bottomrule
\end{tabular}
}
\vspace{-12pt}
\end{table}

\noindent\textbf{Necessity of the Multi-Stage Architecture.}
Finally, we compare the three-stage hierarchical architecture with a single-stage variant, in which all three textual descriptions are concatenated and fused with visual features simultaneously through one cross-attention layer. The single-stage variant achieves an ACC of 65.7\% and a BLEU@4 score of 9.1, both below the performance of the hierarchical design. This result confirms that decomposing semantic guidance into progressive stages is important. By refining the visual representation layer by layer, the proposed model reduces interference among different semantic levels and achieves more robust alignment with multi-level linguistic priors.

\noindent\textbf{Impact of Hierarchical Fusion Order.}
We evaluate all six permutations of textual injection order (Table~\ref{tab:ablation_order}). The proposed coarse-to-fine order, \textbf{Style $\rightarrow$ Subject $\rightarrow$ Detail}, achieves the best overall performance, whereas the reverse order, \textbf{Detail $\rightarrow$ Subject $\rightarrow$ Style}, produces the weakest performance. These results suggest that establishing a global atmospheric context first helps the model interpret subsequent narrative and detail cues more effectively. Reversing this order weakens contextual coherence and degrades both prediction accuracy and generation fluency.

\begin{figure}[t]
    \centering
    \includegraphics[width=0.99\columnwidth]{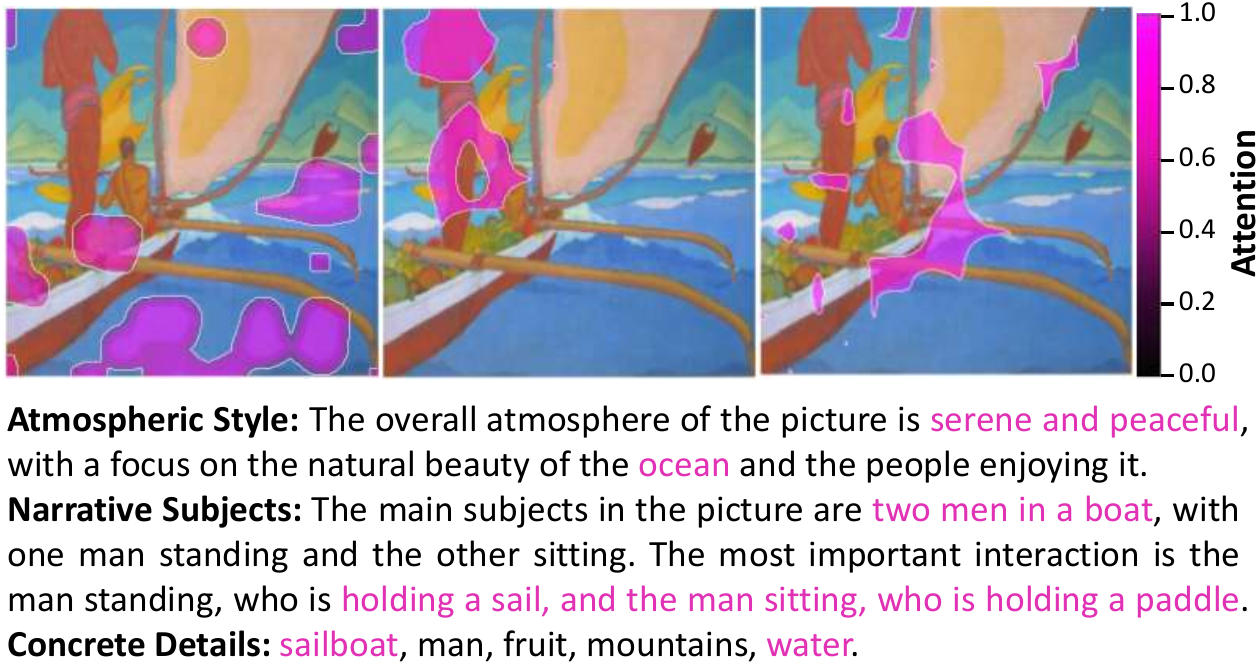}
    \vspace{-5pt}
    \caption{Attention visualization across the three fusion stages. Guided by hierarchical textual priors, the visual focus dynamically aligns with the corresponding semantic regions.}
    \vspace{-5pt}
    \label{fig:attention}
\end{figure}

\begin{figure}[t]
    \centering
    \includegraphics[width=0.99\columnwidth]{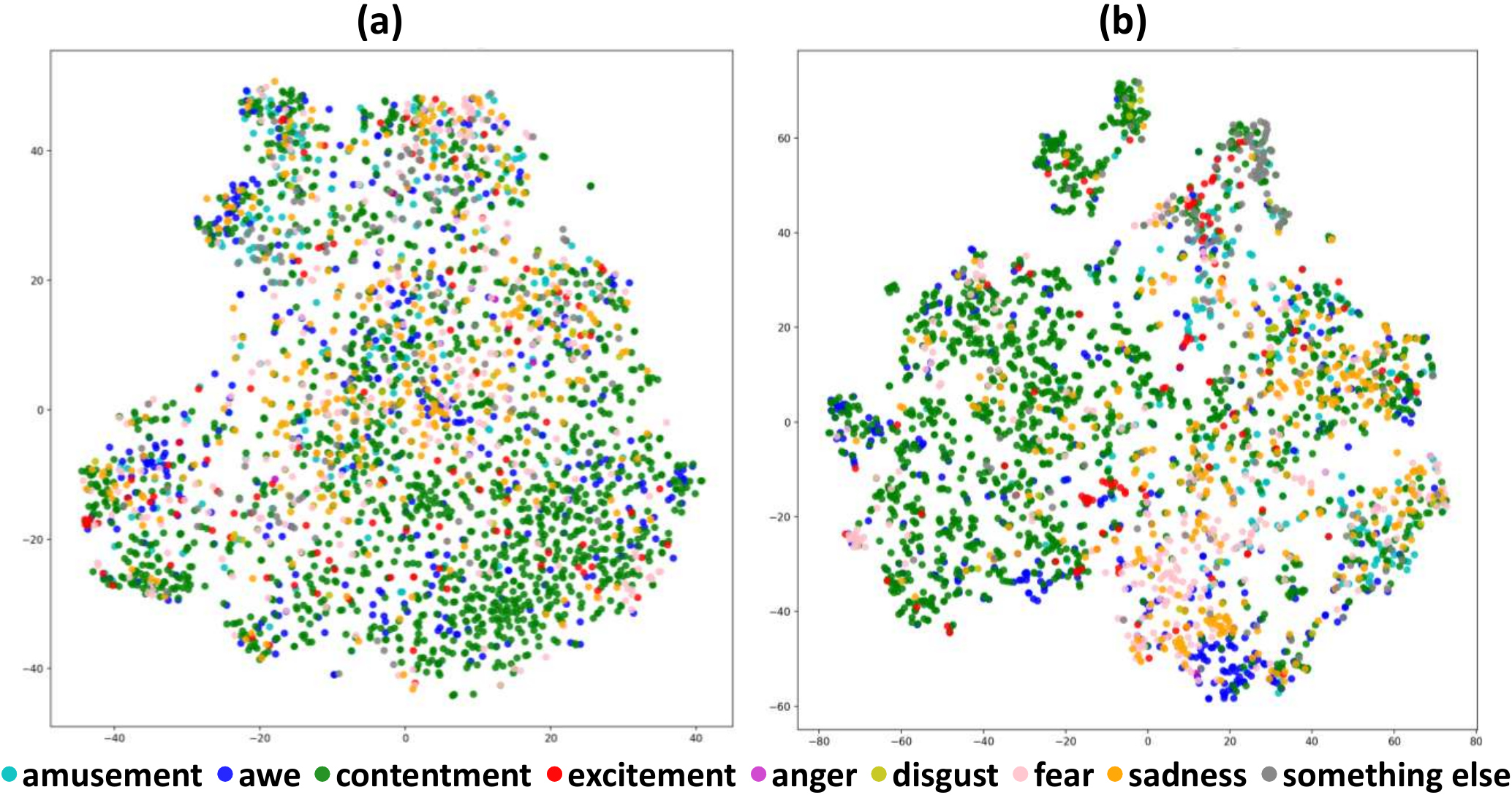}
    \vspace{-7pt}
    \caption{t-SNE visualization of visual feature distributions across nine emotion categories. (a) Implicitly entangled raw visual features from the baseline. (b) Semantically fused features ($\mathbf{V}_{fused}$) by ProFocus.}   
    \vspace{-5pt}
    \label{fig:tsne}
\end{figure}

\begin{figure*}[t]
    \centering
    \includegraphics[width=0.85\textwidth]{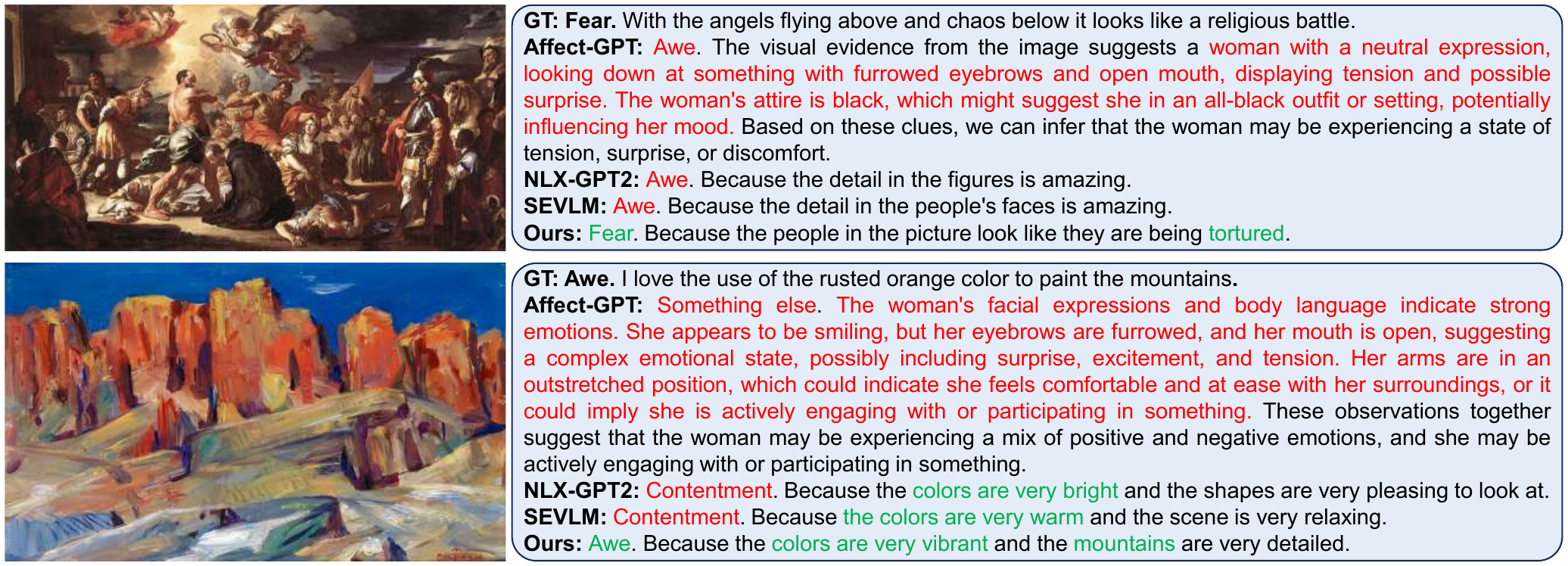}
    \vspace{-7pt}
    \caption{Qualitative comparison of generated explanations on two representative artworks. Red text denotes incorrect emotion predictions or hallucinated content, whereas green text highlights visually grounded cues that support the final explanation.}
    \label{fig:case}
    \vspace{-4pt}
\end{figure*}

\subsection{Qualitative Analysis}
\label{sec:qualitative}

To provide intuitive evidence for the effectiveness of the proposed method, we present several qualitative analyses.

\noindent\textbf{Visualization of Fusion Attention.}
To understand how the Progressive Hint Fusion Module operates, we visualize cross-attention weights in Figure~\ref{fig:attention}. The heatmaps show that attention dynamically shift toward semantically relevant regions under the guidance of progressive textual descriptions. In \textbf{Stage 1}, the text states that ``\textit{The overall atmosphere of the picture is serene and peaceful...}'' and the attention map broadly highlights the background environment, such as the ocean and sky. In \textbf{Stage 2}, the text focuses on ``\textit{two men in a boat... holding a sail... holding a paddle,}'' and the attention immediately shifts to the corresponding human figures and their actions. In \textbf{Stage 3}, guided by the noun list ``\textit{sailboat, man, fruit, mountains, water},'' the attention sharply pinpoints fine-grained objects, especially the pile of fruit on the boat and the edges of the sail. This progressive localization demonstrates that the hierarchical fusion dynamically aligns the visual focus with the corresponding semantic regions.

\noindent\textbf{Visualizing Feature Space Evolution.}
To further verify that ProFocus bridges the affective gap, we employ t-SNE to visualize the feature distributions across nine emotion categories. As shown in Figure~\ref{fig:tsne}(a), the raw visual features from the baseline exhibit a chaotic and entangled distribution with extensive inter-class overlap. By contrast, the semantically fused features $\mathbf{V}_{fused}$ in Figure~\ref{fig:tsne}(b) demonstrate significantly improved intra-class compactness and inter-class separability. This distinct clustering pattern provides strong evidence that the progressive injection of explicit aesthetic priors effectively disentangles the holistic visual feature space.

\noindent\textbf{Comparison of Generated Explanations.}
Figure~\ref{fig:case} presents two representative examples comparing the explanations generated by Affect-GPT, NLX-GPT2, SEVLM, and the proposed model. In the first case, Affect-GPT hallucinates a non-existent female subject, whereas NLX-GPT2 and SEVLM overemphasize generic visual details and incorrectly predict \textit{Awe}. By contrast, the proposed model captures the suffering human figures and produces a fear-related explanation that is consistent with the ground truth. In the second case, Affect-GPT again hallucinates a human subject that is absent from the artwork, while the other baselines provide only generic color-based comments, leading to the incorrect prediction of \textit{Contentment}. The proposed model instead focuses on image-consistent cues, including the vibrant colors and mountain-like forms, and correctly predicts \textit{Awe}. These examples show that hierarchical semantic priors help reduce hallucination and improve both visual grounding and affective reasoning.

\begin{table}[t]
\centering
\caption{Human evaluation results on the ArtEmis v1.0 test set in Mean Opinion Scores on a 1--5 scale. Avg. denotes the average over EG, VR, DR, LF, and OP.}
\vspace{-7pt}
\label{tab:human_eval}
\resizebox{0.7\columnwidth}{!}{%
\begin{tabular}{l|cccccc}
\toprule
\textbf{Method} & \textbf{EG} & \textbf{VR} & \textbf{DR} & \textbf{LF} & \textbf{OP} & \textbf{Avg.} \\ \midrule
NLX-GPT2 & 3.35 & 3.47 & 3.15 & 4.18 & 3.30 & 3.49 \\
SEVLM    & 3.51 & 3.65 & 3.26 & 4.16 & 3.55 & 3.63 \\ \midrule
\textbf{ProFocus} & \textbf{3.91} & \textbf{4.05} & \textbf{3.78} & \textbf{4.20} & \textbf{3.82} & \textbf{3.95} \\ \bottomrule
\end{tabular}
}
\vspace{-20pt}
\end{table}

\subsection{Human Evaluation}
\label{sec:human_eval}

Since automated metrics cannot fully capture semantic alignment and reasoning quality, we conduct a human evaluation. We randomly sample 300 artworks from the ArtEmis v1.0 test set, asking three independent annotators to blindly rate generated explanations on a 1--5 scale. The five criteria include: 
Emotion Grounding (EG), measuring support for the predicted emotion; Visual Relevance (VR), assessing visual fidelity without hallucination; Detail Richness (DR), reflecting the presence of specific fine-grained cues rather than generic templates; Language Fluency (LF), evaluating grammatical naturalness; and Overall Persuasiveness (OP), judging how convincing the rationale is.

As shown in Table~\ref{tab:human_eval}, all models achieve high LF scores due to the strong language prior of the pre-trained GPT decoder. However, ProFocus significantly outperforms baselines across all other metrics. Notably, the substantial gains in DR and VR demonstrate that the explicitly injected progressive hints successfully reduce visual hallucination, enabling the model to generate highly grounded and persuasive affective rationales.

\section{Conclusion}
\label{sec:conclusion}

In this paper, we present ProFocus, a novel framework for interpreting affective experiences in visual art. 
To overcome the limitations of holistic visual representations, which tightly entangle abstract aesthetics and visual metaphors, 
ProFocus models representation learning inspired by the human cognitive pathway.
We introduce a Hierarchical Art Critic (HAC) that leverages a multimodal large language model to decouple artistic visual cues into three structured linguistic priors: atmospheric style, narrative subjects, and concrete details. To seamlessly integrate these multi-level priors, we design a Progressive Hint Fusion (PHF) module that sequentially injects them into visual features in a coarse-to-fine manner. This mechanism mimics human aesthetic appreciation, ensuring that the global atmospheric context properly grounds the interpretation of localized narratives and symbolic evidence. Extensive experiments on the ArtEmis v1.0 and v2.0 datasets demonstrate that ProFocus significantly outperforms existing methods, establishing a new state-of-the-art in both emotion prediction and the generation of visually grounded affective explanations.

\begin{acks}
This work was supported by the National Natural Science Foundation of China (62402471, U22A2094, 62577001), the Yangtze River Delta Science and Technology Innovation Community Joint Research (Basic Research) Project (2025CSJZN01600), and the National Science and Technology Major Project (2025ZD0215301).
\end{acks}

\bibliographystyle{ACM-Reference-Format}
\balance
\bibliography{sample-base}

@InProceedings{Affection,
  author = 	 "Achlioptas, P. and Ovsjanikov, M. and Guibas, L. J. and Tulyakov, S.",
  title =  	 "Affection: Learning Affective Explanations for Real-World Visual Data",
  booktitle = 	 "CVPR",
  pages = 	 "6641-6651",
  year = 	 "2023"
}

@InProceedings{artemisv1,
  author = 	 "Achlioptas, P. and Ovsjanikov, M. and Haydarov, K. and Elhoseiny, M. and Guibas, L. J.",
  title =  	 "ArtEmis: Affective Language for Visual Art",
  booktitle = 	 "CVPR",
  pages = 	 "11569-11579",
  year = 	 "2021"
}

@InProceedings{artemisv2,
  author = 	 "Mohamed, Y. and Khan, F. F. and Haydarov, K. and Elhoseiny, M.",
  title =  	 "It is Okay to Not Be Okay: Overcoming Emotional Bias in Affective Image Captioning by Contrastive Data Collection",
  booktitle = 	 "CVPR",
  pages = 	 "21231-21240",
  year = 	 "2022"
}

@Article{LLaVA-OneVision,
  author = 	 "Li, B. and Zhang, Y. and Guo, D. and Zhang, R. and Li, F. and Zhang, H. and Zhang, K. and Zhang, P. and Li, Y. and Liu, Z. and Li, C.",
  title =  	 "LLaVA-OneVision: Easy Visual Task Transfer",
  journal = 	 "Trans. Mach. Learn. Res.",
  year = 	 "2025"
}

@InProceedings{chen2024internvl,
  author = 	 "Chen, Z. and Wu, J. and Wang, W. and Su, W. and Chen, G. and Xing, S. and Zhong, M. and Zhang, Q. and Zhu, X. and Lu, L. and others",
  title =  	 "Internvl: Scaling up vision foundation models and aligning for generic visual-linguistic tasks",
  booktitle = 	 "CVPR",
  pages = 	 "24185-24198",
  year = 	 "2024"
}

@InProceedings{emotionllama,
  author = 	 "Cheng, Z. and Cheng, Z.-Q. and He, J.-Y. and Wang, K. and Lin, Y. and Lian, Z. and Peng, X. and Hauptmann, A. G.",
  title =  	 "Emotion-LLaMA: Multimodal Emotion Recognition and Reasoning with Instruction Tuning",
  booktitle = 	 "NeurIPS",
  volume={37},
  pages={110805--110853},
  year = 	 "2024"
}

@InProceedings{sevlm,
  author = 	 "Zhang, J. and Zheng, L. and Wang, M. and Guo, D.",
  title =  	 "Training A Small Emotional Vision Language Model for Visual Art Comprehension",
  booktitle = 	 "ECCV",
  volume = 	 "15125",
  pages = 	 "397-413",
  year = 	 "2024"
}

@InProceedings{clip,
  author = 	 "Radford, A. and Kim, J. W. and Hallacy, C. and Ramesh, A. and Goh, G. and Agarwal, S. and Sastry, G. and Askell, A. and Mishkin, P. and Clark, J. and Krueger, G. and Sutskever, I.",
  title =  	 "Learning Transferable Visual Models From Natural Language Supervision",
  booktitle = 	 "ICML",
  volume = 	 "139",
  pages = 	 "8748-8763",
  year = 	 "2021"
}

@Article{gpt2,
  author = 	 "Radford, A. and Wu, J. and Child, R. and Luan, D. and Amodei, D. and Sutskever, I. and et al.",
  title =  	 "Language models are unsupervised multitask learners",
  journal = 	 "OpenAI blog",
  volume = 	 "1",
  number = 	 "8",
  pages = 	 "9",
  year = 	 "2019"
}

@Article{ekman,
  author = 	 "Ekman, P.",
  title =  	 "An argument for basic emotions",
  journal = 	 "Cognition \& emotion",
  volume = 	 "6",
  number = 	 "3-4",
  pages = 	 "169-200",
  year = 	 "1992"
}

@Article{adamw,
  author = 	 "Loshchilov, I. and Hutter, F.",
  title =  	 "Decoupled weight decay regularization",
  journal = 	 "arXiv preprint arXiv:1711.05101",
  year = 	 "2017"
}

@InProceedings{nlxgpt,
  author = 	 "Sammani, F. and Mukherjee, T. and Deligiannis, N.",
  title =  	 "Nlx-gpt: A model for natural language explanations in vision and vision-language tasks",
  booktitle = 	 "CVPR",
  pages = 	 "8322-8332",
  year = 	 "2022"
}

@InProceedings{song2023emotion,
  author = 	 "Song, P. and Guo, D. and Yang, X. and Tang, S. and Yang, E. and Wang, M.",
  title =  	 "Emotion-prior awareness network for emotional video captioning",
  booktitle = 	 "ACM MM",
  pages = 	 "589-600",
  year = 	 "2023"
}

@article{leder2004model,
  title={A model of aesthetic appreciation and aesthetic judgments},
  author={Leder, Helmut and Belke, Benno and Oeberst, Andries and Augustin, Dorothee},
  journal={British journal of psychology},
  volume={95},
  number={4},
  pages={489--508},
  year={2004},
  publisher={Wiley Online Library}
}

@article{hu2024psycollm,
  title={Psycollm: Enhancing llm for psychological understanding and evaluation},
  author={Hu, Jinpeng and Dong, Tengteng and Luo, Gang and Ma, Hui and Zou, Peng and Sun, Xiao and Guo, Dan and Yang, Xun and Wang, Meng},
  journal={IEEE Transactions on Computational Social Systems},
  volume={12},
  number={2},
  pages={539--551},
  year={2024},
  publisher={IEEE}
}

@article{bai2023qwen,
  title={Qwen technical report},
  author={Bai, Jinze and Bai, Shuai and Chu, Yunfei and Cui, Zeyu and Dang, Kai and Deng, Xiaodong and Fan, Yang and Ge, Wenbin and Han, Yu and Huang, Fei and others},
  journal={arXiv preprint arXiv:2309.16609},
  year={2023}
}

@article{achiam2023gpt,
  title={Gpt-4 technical report},
  author={Achiam, Josh and Adler, Steven and Agarwal, Sandhini and Ahmad, Lama and Akkaya, Ilge and Aleman, Florencia Leoni and Almeida, Diogo and Altenschmidt, Janko and Altman, Sam and Anadkat, Shyamal and others},
  journal={arXiv preprint arXiv:2303.08774},
  year={2023}
}

@misc{liu2024llavanext,
    title={LLaVA-NeXT: Improved reasoning, OCR, and world knowledge},
    url={https://llava-vl.github.io/blog/2024-01-30-llava-next/},
    author={Liu, Haotian and Li, Chunyuan and Li, Yuheng and Li, Bo and Zhang, Yuanhan and Shen, Sheng and Lee, Yong Jae},
    month={January},
    year={2024}
}

@inproceedings{mohammad2018obtaining,
  title={Obtaining reliable human ratings of valence, arousal, and dominance for 20,000 English words},
  author={Mohammad, Saif},
  booktitle={Proceedings of the 56th annual meeting of the association for computational linguistics (volume 1: Long papers)},
  pages={174--184},
  year={2018}
}

@article{lian2025affectgpt,
  title={Affectgpt: A new dataset, model, and benchmark for emotion understanding with multimodal large language models},
  author={Lian, Zheng and Chen, Haoyu and Chen, Lan and Sun, Haiyang and Sun, Licai and Ren, Yong and Cheng, Zebang and Liu, Bin and Liu, Rui and Peng, Xiaojiang and others},
  journal={arXiv preprint arXiv:2501.16566},
  year={2025}
}

@article{zhang2025videmo,
  title={VidEmo: Affective-Tree Reasoning for Emotion-Centric Video Foundation Models},
  author={Zhang, Zhicheng and Wang, Weicheng and Zhu, Yongjie and Qin, Wenyu and Wan, Pengfei and Zhang, Di and Yang, Jufeng},
  journal={arXiv preprint arXiv:2511.02712},
  year={2025}
}

@inproceedings{xie2024emovit,
  title={Emovit: Revolutionizing emotion insights with visual instruction tuning},
  author={Xie, Hongxia and Peng, Chu-Jun and Tseng, Yu-Wen and Chen, Hung-Jen and Hsu, Chan-Feng and Shuai, Hong-Han and Cheng, Wen-Huang},
  booktitle={Proceedings of the IEEE/CVF Conference on Computer Vision and Pattern Recognition},
  pages={26596--26605},
  year={2024}
}

@book{bell1916art,
  title={Art},
  author={Bell, Clive},
  year={1916},
  publisher={Chatto \& Windus}
}

@inproceedings{garcia2018read,
  title={How to read paintings: semantic art understanding with multi-modal retrieval},
  author={Garcia, Noa and Vogiatzis, George},
  booktitle={Proceedings of the European Conference on Computer Vision (ECCV) Workshops},
  pages={0--0},
  year={2018}
}

@book{picard2000affective,
  title={Affective computing},
  author={Picard, Rosalind W},
  year={2000},
  publisher={MIT press}
}

@article{chatterjee2014neuroaesthetics,
  title={Neuroaesthetics},
  author={Chatterjee, Anjan and Vartanian, Oshin},
  journal={Trends in cognitive sciences},
  volume={18},
  number={7},
  pages={370--375},
  year={2014},
  publisher={Elsevier}
}

@article{comanici2025gemini,
  title={Gemini 2.5: Pushing the frontier with advanced reasoning, multimodality, long context, and next generation agentic capabilities},
  author={Comanici, Gheorghe and Bieber, Eric and Schaekermann, Mike and Pasupat, Ice and Sachdeva, Noveen and Dhillon, Inderjit and Blistein, Marcel and Ram, Ori and Zhang, Dan and Rosen, Evan and others},
  journal={arXiv preprint arXiv:2507.06261},
  year={2025}
}

@inproceedings{lin2014microsoft,
  title={Microsoft coco: Common objects in context},
  author={Lin, Tsung-Yi and Maire, Michael and Belongie, Serge and Hays, James and Perona, Pietro and Ramanan, Deva and Doll{\'a}r, Piotr and Zitnick, C Lawrence},
  booktitle={European conference on computer vision},
  pages={740--755},
  year={2014},
  organization={Springer}
}

@article{gatys2015neural,
  title={A neural algorithm of artistic style},
  author={Gatys, Leon A and Ecker, Alexander S and Bethge, Matthias},
  journal={arXiv preprint arXiv:1508.06576},
  year={2015}
}

@inproceedings{mazzone2019art,
  title={Art, creativity, and the potential of artificial intelligence},
  author={Mazzone, Marian and Elgammal, Ahmed},
  booktitle={Arts},
  volume={8},
  number={1},
  pages={26},
  year={2019},
  organization={MDPI}
}

@article{yang2021stimuli,
  title={Stimuli-aware visual emotion analysis},
  author={Yang, Jingyuan and Li, Jie and Wang, Xiumei and Ding, Yuxuan and Gao, Xinbo},
  journal={IEEE Transactions on Image Processing},
  volume={30},
  pages={7432--7445},
  year={2021},
  publisher={IEEE}
}

@inproceedings{danvevcek2022emoca,
  title={Emoca: Emotion driven monocular face capture and animation},
  author={Dan{\v{e}}{\v{c}}ek, Radek and Black, Michael J and Bolkart, Timo},
  booktitle={Proceedings of the IEEE/CVF conference on computer vision and pattern recognition},
  pages={20311--20322},
  year={2022}
}

@inproceedings{yang2018weakly,
  title={Weakly supervised coupled networks for visual sentiment analysis},
  author={Yang, Jufeng and She, Dongyu and Lai, Yu-Kun and Rosin, Paul L and Yang, Ming-Hsuan},
  booktitle={Proceedings of the IEEE conference on computer vision and pattern recognition},
  pages={7584--7592},
  year={2018}
}

@article{krumhuber2023role,
  title={The role of facial movements in emotion recognition},
  author={Krumhuber, Eva G and Skora, Lina I and Hill, Harold CH and Lander, Karen},
  journal={Nature Reviews Psychology},
  volume={2},
  number={5},
  pages={283--296},
  year={2023},
  publisher={Nature Publishing Group US New York}
}

@article{zeng2025glm,
  title={Glm-4.5: Agentic, reasoning, and coding (arc) foundation models},
  author={Zeng, Aohan and Lv, Xin and Zheng, Qinkai and Hou, Zhenyu and Chen, Bin and Xie, Chengxing and Wang, Cunxiang and Yin, Da and Zeng, Hao and Zhang, Jiajie and others},
  journal={arXiv preprint arXiv:2508.06471},
  year={2025}
}

@inproceedings{yang2023emoset,
  title={Emoset: A large-scale visual emotion dataset with rich attributes},
  author={Yang, Jingyuan and Huang, Qirui and Ding, Tingting and Lischinski, Dani and Cohen-Or, Danny and Huang, Hui},
  booktitle={Proceedings of the IEEE/CVF International Conference on Computer Vision},
  pages={20383--20394},
  year={2023}
}

@inproceedings{xu2022mdan,
  title={Mdan: Multi-level dependent attention network for visual emotion analysis},
  author={Xu, Liwen and Wang, Zhengtao and Wu, Bin and Lui, Simon},
  booktitle={Proceedings of the IEEE/CVF Conference on Computer Vision and Pattern Recognition},
  pages={9479--9488},
  year={2022}
}

@book{barr2019cubism,
  title={Cubism and abstract art},
  author={Barr Jr, Alfred H},
  year={2019},
  publisher={Routledge}
}

@article{o2015introduction,
  title={An introduction to convolutional neural networks},
  author={O'shea, Keiron and Nash, Ryan},
  journal={arXiv preprint arXiv:1511.08458},
  year={2015}
}

@article{seo2004role,
  title={The role of affective experience in work motivation},
  author={Seo, Myeong-Gu and Barrett, Lisa Feldman and Bartunek, Jean M},
  journal={Academy of management review},
  volume={29},
  number={3},
  pages={423--439},
  year={2004},
  publisher={Academy of management Briarcliff Manor, NY 10510}
}

@article{fernandez2021affective,
  title={Affective experience in the predictive mind: a review and new integrative account},
  author={Fernandez Velasco, Pablo and Loev, Slawa},
  journal={Synthese},
  volume={198},
  number={11},
  pages={10847--10882},
  year={2021},
  publisher={Springer}
}

@inproceedings{wu2025enriching,
  title={Enriching multimodal sentiment analysis through textual emotional descriptions of visual-audio content},
  author={Wu, Sheng and He, Dongxiao and Wang, Xiaobao and Wang, Longbiao and Dang, Jianwu},
  booktitle={Proceedings of the AAAI Conference on Artificial Intelligence},
  volume={39},
  number={2},
  pages={1601--1609},
  year={2025}
}

@article{wu2025comprehensive,
  title={A comprehensive review of multimodal emotion recognition: Techniques, challenges, and future directions},
  author={Wu, You and Mi, Qingwei and Gao, Tianhan},
  journal={Biomimetics},
  volume={10},
  number={7},
  pages={418},
  year={2025},
  publisher={MDPI}
}

@article{wei2022chain,
  title={Chain-of-thought prompting elicits reasoning in large language models},
  author={Wei, Jason and Wang, Xuezhi and Schuurmans, Dale and Bosma, Maarten and Xia, Fei and Chi, Ed and Le, Quoc V and Zhou, Denny and others},
  journal={Advances in neural information processing systems},
  volume={35},
  pages={24824--24837},
  year={2022}
}

@article{shao2024visual,
  title={Visual cot: Advancing multi-modal language models with a comprehensive dataset and benchmark for chain-of-thought reasoning},
  author={Shao, Hao and Qian, Shengju and Xiao, Han and Song, Guanglu and Zong, Zhuofan and Wang, Letian and Liu, Yu and Li, Hongsheng},
  journal={Advances in Neural Information Processing Systems},
  volume={37},
  pages={8612--8642},
  year={2024}
}

@inproceedings{zhang2026cmmcot,
  title={Cmmcot: Enhancing complex multi-image comprehension via multi-modal chain-of-thought and memory augmentation},
  author={Zhang, Guanghao and Zhong, Tao and Xia, Yan and Liu, Mushui and Yu, Zhelun and Li, Haoyuan and He, Wanggui and She, Dong and Wang, Yi and Jiang, Hao},
  booktitle={Proceedings of the AAAI Conference on Artificial Intelligence},
  volume={40},
  number={15},
  pages={12430--12438},
  year={2026}
}

@article{zhang2026affective,
  title={Affective computing in the era of large language models: A survey from the nlp perspective},
  author={Zhang, Yiqun and Yang, Xiaocui and Xu, Xingle and Gao, Zeran and Huang, Yijie and Mu, Shiyi and Feng, Shi and Wang, Daling and Zhang, Yifei and Song, Kaisong and others},
  journal={Knowledge-Based Systems},
  pages={115411},
  year={2026},
  publisher={Elsevier}
}

@misc{cheng2026mojitomodaljointlearning,
      title={MOJITO: Modal Joint Learning for Unified End-to-End Autonomous Driving}, 
      author={Zhijing Cheng and Xuancheng Zhang and Donglin Di and Lei Fan and Baorui Ma and Hao Li and Xun Yang},
      year={2026},
      eprint={2607.23511},
      archivePrefix={arXiv},
      primaryClass={cs.CV},
      url={https://arxiv.org/abs/2607.23511}, 
}

@inproceedings{gooch2002artistic,
  title={Artistic vision: painterly rendering using computer vision techniques},
  author={Gooch, Bruce and Coombe, Greg and Shirley, Peter},
  booktitle={Proceedings of the 2nd international symposium on Non-photorealistic animation and rendering},
  pages={83--ff},
  year={2002}
}

@article{cai2015cross,
  title={The cross-depiction problem: Computer vision algorithms for recognising objects in artwork and in photographs},
  author={Cai, Hongping and Wu, Qi and Corradi, Tadeo and Hall, Peter},
  journal={arXiv preprint arXiv:1505.00110},
  year={2015}
}

@article{wu2025review,
  title={A Review on Research and Application of Al-based Image Analysis in the field of Computer Vision},
  author={Wu, Pan and He, Xiaoqiang and Dai, Wenhao and Zhou, Jingwei and Shang, Yutong and Fan, Yourong and Hu, Tao},
  journal={IEEE Access},
  year={2025},
  publisher={IEEE}
}

@article{song2026bridging,
  title={Bridging Subjectivity in Affective Explanation Captioning via Consensus-Prompted Emotion Reasoning},
  author={Song, Peipei and Zhang, Zhiyan and Chen, Weidong and Hu, Jinpeng and Yang, Xun and Chang, Xiaojun},
  journal={IEEE Transactions on Image Processing},
  year={2026},
  publisher={IEEE}
}

@inproceedings{han2026omni,
  title={Omni-Perception Policy Optimization for Multimodal Emotion Reasoning},
  author={Han, Zhiyuan and Zhu, Beier and Tong, Wenwen and Shao, Pengyang and Song, Peipei and Wang, Xinyi and Chen, Jiangnan and Lu, Lewei and Yang, Xun},
  booktitle={Forty-third International Conference on Machine Learning},
  year={2026}
}

@article{zhang2026benchmarking,
  title={Benchmarking Dynamic Affective Reasoning: A Viewer-Centric Video Emotion Dataset},
  author={Zhang, Zhiyan and Song, Peipei and Hu, Jinpeng and Jia, Jingyang and Yang, Xun and Chang, Xiaojun},
  journal={arXiv preprint arXiv:2607.10238},
  year={2026}
}

@inproceedings{han2025benchmarking,
  title={Benchmarking and bridging emotion conflicts for multimodal emotion reasoning},
  author={Han, Zhiyuan and Zhu, Beier and Xu, Yanlong and Song, Peipei and Yang, Xun},
  booktitle={Proceedings of the 33rd ACM International Conference on Multimedia},
  pages={5528--5537},
  year={2025}
}

@inproceedings{hu2025beyond,
  title={Beyond emotion recognition: A multi-turn multimodal emotion understanding and reasoning benchmark},
  author={Hu, Jinpeng and Shi, Hongchang and Dai, Chongyuan and Li, Zhuo and Song, Peipei and Wang, Meng},
  booktitle={Proceedings of the 33rd ACM International Conference on Multimedia},
  pages={5814--5823},
  year={2025}
}

@article{song2024emotional,
  title={Emotional video captioning with vision-based emotion interpretation network},
  author={Song, Peipei and Guo, Dan and Yang, Xun and Tang, Shengeng and Wang, Meng},
  journal={IEEE Transactions on Image Processing},
  volume={33},
  pages={1122--1135},
  year={2024},
  publisher={IEEE}
}

@inproceedings{zhang2026stimuli,
  title={Stimuli-aware emotion adaptor for enhancing llm in affective explanation captioning},
  author={Zhang, Zhiyan and Song, Peipei and Hu, Jinpeng and Chen, Weidong and Ni, Lin and Yang, Xun},
  booktitle={ICASSP 2026-2026 IEEE International Conference on Acoustics, Speech and Signal Processing (ICASSP)},
  pages={10662--10666},
  year={2026},
  organization={IEEE}
}

\clearpage
\appendix

\section{Baseline Details}
\label{sec:app_baseline}

To comprehensively evaluate the proposed ProFocus framework, we compare it against multiple strong baselines, ranging from task-specific architectures trained on the ArtEmis datasets to recent Multimodal Large Language Models (MLLMs) tailored for affective computing. The implementation details of these baselines are summarized below:

\noindent\textbf{M2 \cite{artemisv1}.} This is a Meshed-Memory Transformer model adopted in the original ArtEmis benchmark for affective image captioning. Specifically, it employs Faster R-CNN to extract bottom-up visual features from the input image. These features are then fed into a transformer-based encoder-decoder architecture to generate emotion-grounded textual explanations.
    
\noindent\textbf{SAT \cite{artemisv1}.} This is an adaptation of the classic Show-Attend-Tell architecture for visual art emotion explanation. It combines an image encoder with a LSTM decoder to generate affective explanations. It utilizes a standard Convolutional Neural Network (e.g., ResNet) as the image encoder to extract spatial features. Subsequently, a word- and image-attentive LSTM decoder is employed to dynamically focus on relevant visual regions while generating the affective rationales.
    
\noindent\textbf{NLX-GPT2 \cite{nlxgpt}.} This is a natural language explanation framework that jointly predicts an answer and its explanation as a unified text generation task. It combines a visual encoder with a Distilled GPT-2 decoder and generates the prediction and corresponding explanation autoregressively.
    
\noindent\textbf{SEVLM \cite{sevlm}.} This is an efficient emotional vision-language model tailored specifically for artistic images, which serves as our direct baseline (as comprehensively detailed in Section 3.1). It combines a frozen CLIP visual encoder with a GPT-2 decoder. To retain interpretable emotion analysis, the model is optimized via a multi-task objective that incorporates VAD-consistency and contrastive alignment losses.
    
\noindent\textbf{Emotion-LLaMA \cite{emotionllama}.} This is a multimodal large language model designed for multimodal emotion recognition and reasoning. Specifically, it leverages a hybrid multi-view visual encoder combining MAE, VideoMAE, and EVA to capture comprehensive visual cues. These representations are then processed by a LLaMA2-chat decoder optimized via parameter-efficient LoRA tuning. For static image evaluation, we adapt the model by processing a single visual frame as input.
    
\noindent\textbf{AffectGPT \cite{lian2025affectgpt}.} This is a comprehensive MLLM tailored for complex emotion understanding. It employs a CLIP ViT-L as the visual encoder and Qwen2.5 as the core language model. Notably, it introduces a pre-fusion operation (e.g., a Q-Former or attention-based mechanism) outside the LLM to explicitly enhance multimodal feature integration. We evaluate its adapted reasoning capability on static artworks.
    
\noindent\textbf{Videmo \cite{zhang2025videmo}.} This is a cutting-edge video emotion foundation model built upon the Qwen2.5-VL architecture. It is uniquely trained through curriculum emotion learning and an innovative affective-tree reinforcement learning strategy (GRPO) to enhance structural emotional reasoning. In our experiments, we evaluate its zero-shot and transferred affective reasoning capabilities on the static ArtEmis images.

\section{Model Complexity and Inference Efficiency}
\label{sec:app_implementation}

\begin{table}[H]
\centering
\caption{Summary of model complexity, hardware resources, and inference efficiency.}
\label{tab:app_complexity}
\resizebox{0.9\columnwidth}{!}{%
\begin{tabular}{lc}
\toprule
\textbf{Configuration / Metric} & \textbf{Value} \\ \midrule
\textbf{Hardware Setup} & \\
\quad - Training & 2 $\times$ NVIDIA Tesla V100 (32GB) \\
\quad - Inference & 1 $\times$ NVIDIA Tesla V100 (32GB) \\ \midrule
\textbf{Total Model Parameters} & \textbf{$\sim$212.0 M} \\
\quad - Frozen CLIP ViT-B/16 Encoder & 86.2 M \\
\quad - Custom GPT-2 Decoder (6-layer) & 100.2 M \\
\quad - Progressive Hint Fusion Module & $\sim$26.0 M \\ \midrule
\textbf{Computational Complexity} & \textbf{22.83 GFLOPs} \\ \midrule
\textbf{Average Inference Speed} & \textbf{13.78 FPS} \\ \bottomrule
\end{tabular}%
}
\end{table}

To provide a comprehensive overview of the computational overhead and practical efficiency of our proposed framework, we detail the model complexity, hardware resources, and decoding configurations in Table~\ref{tab:app_complexity}. 

As shown, ProFocus maintains a reasonable parameter scale of approximately 212M. Thanks to the frozen visual encoder and the lightweight 6-layer GPT-2 decoder design, the framework requires 22.83 GFLOPs per full forward pass. The entire affective reasoning pipeline (excluding the offline LLaVA extraction) is trained on two NVIDIA Tesla V100-SXM2-32GB GPUs. To evaluate the inference efficiency, we generate explanations on a single V100 GPU using standard autoregressive sampling (\texttt{temperature}=1.0, \texttt{top\_p}=0.9, \texttt{top\_k}=0) with a maximum sequence length capped at 25. Under these specific conditions, the model achieves a highly acceptable inference speed of 13.78 frames per second (FPS), demonstrating its efficiency for image-level affective interpreting.

\begin{figure*}[t]
    \centering
    \begin{subfigure}[t]{0.49\textwidth}
        \centering
        \includegraphics[width=\linewidth]{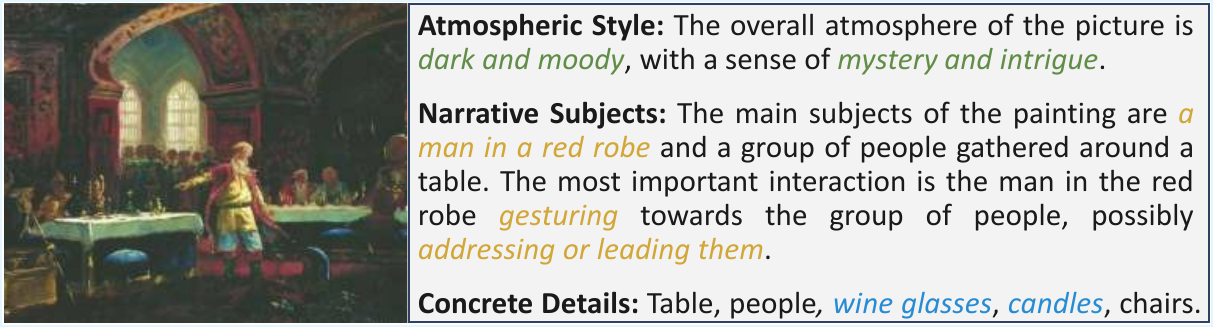}
        \caption{}
    \end{subfigure}
    \hfill
    \begin{subfigure}[t]{0.49\textwidth}
        \centering
        \includegraphics[width=\linewidth]{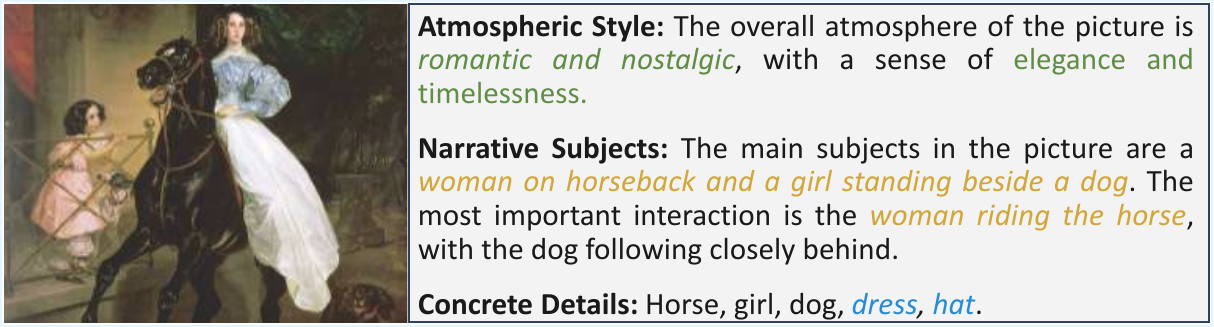}
        \caption{}
    \end{subfigure}
    \vspace{-8pt}
    \caption{Qualitative examples of the hierarchical semantic priors extracted by the Hierarchical Art Critic. For each representative artwork, the implicitly entangled visual aesthetics are explicitly decoupled into structured linguistic cues across three cognitive levels.}
    \label{fig:app_llava_examples}
    \vspace{-4pt}
\end{figure*}

\section{Prompt for Hierarchical Art Critic}
\label{sec:app_llava}

To explicitly decouple abstract visual aesthetics into three hierarchical granularities, we use LLaVA-1.6 as our Hierarchical Art Critic and apply the following prompts:

\vspace{1ex}
\noindent \textbf{Atmospheric Style (Stage 1):} \\
\textit{Describe the overall artistic style and the dominant color palette of this artwork in under 60 words. Focus only on visual style (such as level of abstraction, line quality, texture, and compositional approach) and the main colors and their relative prominence. Do not mention emotions, mood, story, or specific objects or figures.}

\vspace{1ex}
\noindent \textbf{Narrative Subjects (Stage 2):} \\
\textit{In under 60 words, name the primary objects or figures in the image and summarize their main spatial relationship or interaction using only factual visual details. Focus on positions, poses, and observable actions. Do not mention emotions, mood, atmosphere, or artistic style and colors.}

\vspace{1ex}
\noindent \textbf{Concrete Details (Stage 3):} \\
\textit{List 5 to 10 distinct concrete nouns that capture additional detailed elements visible in the artwork, using only a comma-separated list. Use single nouns only, with no adjectives or phrases. Focus on secondary or background elements that you have NOT mentioned in your previous answers. Do not include any emotions, mood, atmosphere, or interpretations, and keep the whole answer under 60 words.}
\vspace{1ex}

Even given the highly subjective nature of visual art, LLaVA-1.6 can provide robust and structurally consistent aesthetic priors, primarily due to its powerful zero-shot descriptive capability. However, a small portion of the raw outputs may slightly deviate from the strict formatting instructions (e.g., occasionally generating short descriptive phrases instead of strictly comma-separated nouns, or slightly exceeding the 60-word limit). We preserve these minor formatting occurrences without manual deletion, as their low frequency does not disrupt the training distribution and naturally reflects the inherent difficulty of rigidly standardizing abstract artistic interpretations. Furthermore, outputs that severely exceeded the length limit and were forcefully truncated during batch generation were filtered and re-processed via single-image inference to ensure structural integrity.

Figure~\ref{fig:app_llava_examples} illustrates the extracted three-layer textual priors for representative artworks. As shown, the HAC effectively distills global atmosphere, structured narratives, and symbolic nouns into explicit linguistic cues.

\section{Robustness of HAC Priors}
\label{sec:app_hac_robustness}

\begin{table}[H]
\centering
\caption{Sensitivity to the MLLM used for generating HAC priors on ArtEmis v1.0.}
\vspace{-4pt}
\label{tab:app_critic_swap}
\resizebox{0.99\columnwidth}{!}{%
\begin{tabular}{lccccccc}
\toprule
HAC prior source & ACC & B@1 & B@2 & B@3 & B@4 & M & R \\
\midrule
InternVL-2.5-8B & 65.9 & 54.8 & 30.7 & 16.7 & 9.4 & 14.0 & 30.5 \\
Qwen3.5-9B & \textbf{66.5} & 55.1 & \textbf{31.2} & 16.9 & 9.7 & 14.0 & 30.6 \\
\textbf{ProFocus (LLaVA-1.6-7B)} & 66.1 & \textbf{55.3} & \textbf{31.2} & \textbf{17.1} & \textbf{9.8} & \textbf{14.2} & \textbf{30.8} \\
\bottomrule
\end{tabular}}
\vspace{-4pt}
\end{table}

To examine whether ProFocus depends on a specific MLLM for generating hierarchical priors, we replace LLaVA-1.6-7B with InternVL-2.5-8B and Qwen3.5-9B. As shown in Table~\ref{tab:app_critic_swap}, both emotion recognition and explanation performance remain stable across different critics, suggesting that the effectiveness of ProFocus is not tied to a specific MLLM.

We further analyze the quality of the generated priors through cross-critic inconsistency and manual verification. The observed failures mainly arise from object misidentification or ambiguous interpretations of abstract artworks. The corresponding failure rates are 0.10\%, 0.04\%, and 0.16\% for InternVL, LLaVA, and Qwen, respectively.

To further assess robustness to erroneous priors, we evaluate ProFocus on ArtEmis v1.0 samples with hallucinated LLaVA outputs. The model achieves ACC/\allowbreak B@1/\allowbreak B@2/\allowbreak B@3/\allowbreak B@4/\allowbreak M/\allowbreak R of 61.8/\allowbreak 56.4/\allowbreak 32.4/\allowbreak 17.0/\allowbreak 8.0/\allowbreak 14.7/\allowbreak 31.4. Although incorrect priors degrade emotion recognition, the model retains meaningful explanation performance, indicating partial robustness to imperfect critic outputs.

\section{Additional Ablation Studies}
\label{sec:app_ablation}

\begin{table}[H]
\centering
\caption{Ablation study on pairwise combinations of semantic priors. We omit one specific prior stage to evaluate the contribution of the remaining two.}
\vspace{-4pt}
\label{tab:app_pairwise}
\resizebox{\columnwidth}{!}{
\begin{tabular}{l|ccccccc}
\toprule
\textbf{Retained Stages} & \textbf{ACC} & \textbf{B@1} & \textbf{B@2} & \textbf{B@3} & \textbf{B@4} & \textbf{M} & \textbf{R} \\ \midrule
Style + Subject (w/o Detail) & 64.0 & 54.3 & 30.3 & 16.5 & 9.3 & 13.5 & 30.5 \\
Detail + Style (w/o Subject) & 65.4 & 54.1 & 30.3 & 16.4 & 9.1 & 13.8 & 30.3 \\
Subject + Detail (w/o Style) & 64.0 & 54.3 & 30.4 & 16.4 & 9.2 & 13.7 & 30.4 \\ \midrule
\textbf{Full (Style + Subject + Detail)} & \textbf{66.1} & \textbf{55.3} & \textbf{31.2} & \textbf{17.1} & \textbf{9.8} & \textbf{14.2} & \textbf{30.8} \\ \bottomrule
\end{tabular}
}
\vspace{-4pt}
\end{table}

To further validate our design choices, we conduct additional ablation experiments on the ArtEmis v1.0 dataset focusing on pairwise semantic combinations. For all experiments, we report the checkpoint results optimized for the best overall generation quality. While we demonstrated in the main paper that omitting the multi-stage fusion entirely degrades performance, Table~\ref{tab:app_pairwise} presents a more fine-grained analysis by ablating exactly one semantic stage at a time (i.e., retaining only two out of the three priors). The results clearly show that removing any single level of cognitive understanding—whether it be the global style, the subject interaction, or the localized details—causes a noticeable drop in both emotion recognition accuracy and explanation fluency. This further corroborates that the three hierarchical stages are strictly mutually complementary.

\begin{figure*}[t]
    \centering
    \includegraphics[width=0.99\textwidth]{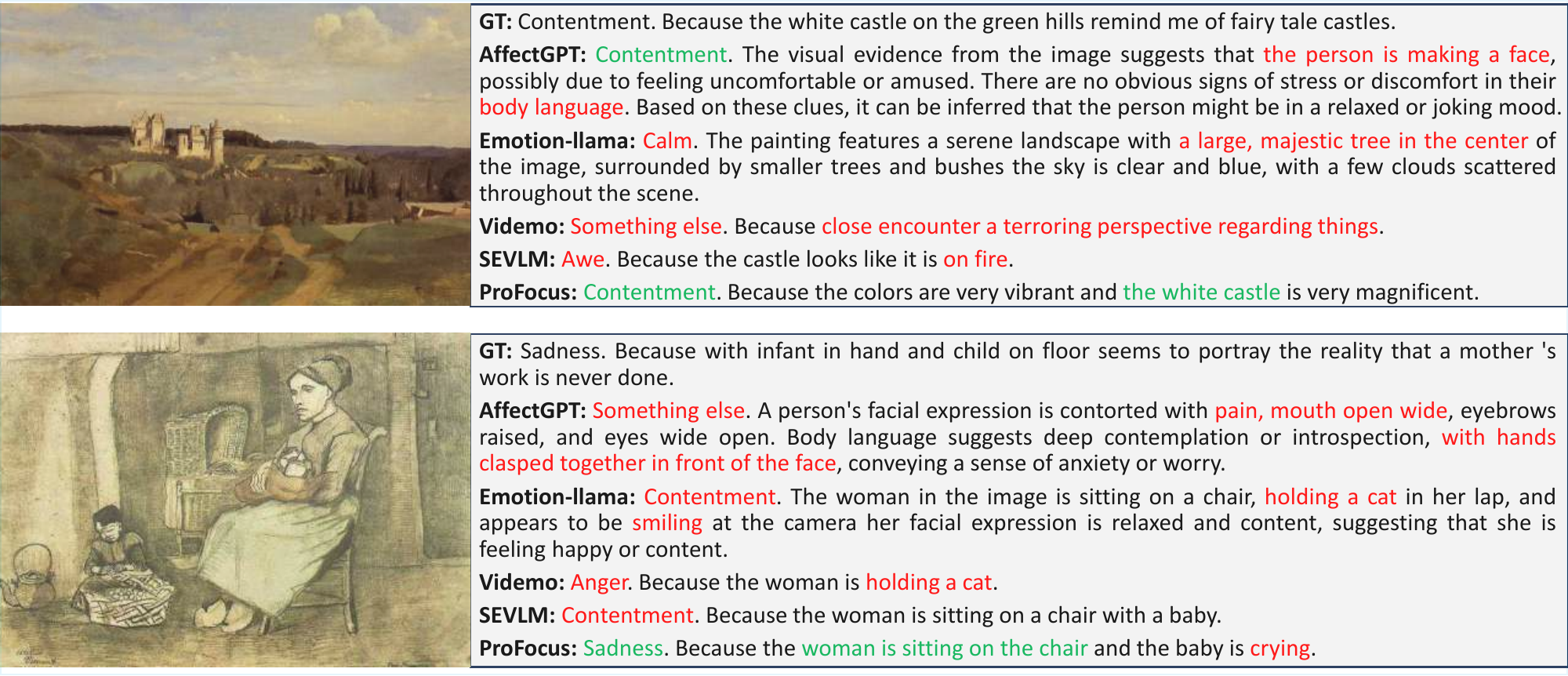}
    \vspace{-4pt}
    \caption{Qualitative comparison of generated explanations. \textbf{Top}: A success case where ProFocus accurately captures the aesthetic elements without hallucination. \textbf{Bottom}: A failure case where ProFocus predicts the correct emotion and subject, but hallucinates a specific action (``crying'') to justify the abstract sadness.}
    \label{fig:app_qualitative_cases}
    \vspace{-4pt}
\end{figure*}

\section{Hyperparameter Sensitivity Analysis}
\label{sec:app_hyperparams}

In this section, we conduct sensitivity analyses on three key hyperparameters of the proposed framework: the number of cross-attention heads, the multi-task loss weighting coefficients ($\lambda_{lm}:\lambda_{cl}:\lambda_{vad}$), and the depth of each fusion module (i.e., the number of Transformer layers per stage). 

\begin{table}[H]
\centering
\caption{Hyperparameter sensitivity experiments on attention heads, loss coefficients, and fusion layer depth. The last row denotes the default configuration of ProFocus.}
\vspace{-8pt}
\label{tab:app_hyperparams}
\resizebox{\columnwidth}{!}{
\begin{tabular}{l|l|ccccccc}
\toprule
\textbf{Hyperparameter} & \textbf{Variant} & \textbf{ACC} & \textbf{B@1} & \textbf{B@2} & \textbf{B@3} & \textbf{B@4} & \textbf{M} & \textbf{R} \\ \midrule
\multirow{2}{*}{\begin{tabular}[c]{@{}l@{}}Cross-Attention Heads\end{tabular}} 
& Heads = 8  & 64.5 & 54.7 & 30.7 & 16.6 & 9.3 & 13.8 & 30.7 \\
& Heads = 16 & 64.2 & 54.5 & 30.6 & 16.2 & 9.3 & 14.0 & 30.7 \\
\midrule
\multirow{2}{*}{\begin{tabular}[c]{@{}l@{}}Loss Weights \end{tabular}} 
& 1.5 : 1.0 : 1.0 & 64.3 & 54.1 & 30.1 & 16.3 & 9.2 & 14.0 & 30.4 \\
& 1.0 : 1.0 : 1.5 & 64.6 & 54.5 & 30.4 & 16.7 & 9.5 & 14.1 & 30.5 \\
\midrule
\multirow{2}{*}{\begin{tabular}[c]{@{}l@{}}Fusion Layers \end{tabular}} 
& 2 Layers & \textbf{66.1} & 54.5 & 30.6 & 16.8 & 9.5 & 14.0 & 30.6 \\
& 3 Layers & 65.9 & 54.7 & 30.6 & 16.6 & 9.1 & 13.8 & 30.4 \\ \midrule
\multicolumn{2}{l|}{\textbf{Ours (Default Settings)}} & \textbf{66.1} & \textbf{55.3} & \textbf{31.2} & \textbf{17.1} & \textbf{9.8} & \textbf{14.2} & \textbf{30.8} \\ \bottomrule
\end{tabular}%
}
\vspace{-4pt}
\end{table}

As shown in Table~\ref{tab:app_hyperparams}, we compare various hyperparameter mutations against our \textbf{default settings} (12 heads, 1.0:1.0:1.0 loss weights, and 1 fusion layer, shown in the last row). 

Setting the cross-attention heads to 12 yields the best balance. Modifying the number of heads to 8 or 16 restricts the representational capacity or introduces unnecessary attention noise, leading to noticeable performance drops across all metrics.

For the multi-task objective, shifting the default equal balance (1.0:1.0:1.0) to heavily penalize language modeling (1.5:1.0:1.0) degrades explanation fluency, dropping the BLEU@4 score from 9.8 to 9.2. Similarly, heavily weighting VAD consistency (1.0:1.0:1.5) drops the BLEU@4 score to 9.5. This confirms that the equal contribution of the three losses provides the most stable optimization gradient for both generation and classification tasks.

For the fusion depth, our default setting uses 1 Transformer layer (one cross-attention block followed by a feed-forward network) for each progressive stage. Interestingly, increasing this depth to 2 layers maintains the peak emotion recognition accuracy (66.1\%), but causes a degradation in explanation generation (BLEU@4 drops from 9.8 to 9.5). Further increasing the depth to 3 layers degrades both the accuracy (65.9\%) and the explanation quality. This indicates that a single layer is entirely sufficient for the visual features to absorb specific linguistic hints. Stacking multiple layers within a single semantic stage does not provide additional classification benefits and instead introduces redundant parameters that interfere with the semantic alignment of the language decoder.

\section{Qualitative Results}
\label{sec:app_qualitative}

To provide a deeper understanding of the generation quality and the limitations of the proposed framework, we present additional qualitative results, comparing ProFocus with state-of-the-art MLLMs and task-specific baselines. Figure~\ref{fig:app_qualitative_cases} illustrates both a success case and a typical failure case. 

\noindent\textbf{Success Cases.} 
In the first example (top of Figure~\ref{fig:app_qualitative_cases}), the artwork depicts a serene landscape featuring a prominent castle on a hill. General-purpose MLLMs severely struggle with this artistic image: AffectGPT completely hallucinates a human subject (``the person is making a face'', ``body language''), Emotion-llama predicts an out-of-domain label (``Calm'') and hallucinates a ``giant tree'', while Videmo produces nonsensical text. The baseline SEVLM incorrectly predicts ``Awe'' and misinterprets the lighting, claiming the castle is ``on fire''. 
In contrast, ProFocus accurately predicts ``Contentment'' and grounds its explanation in factual aesthetic evidence (``the white castle'' and ``vibrant colors''). By leveraging the explicitly decoupled semantic priors, our model effectively avoids the generic templates and severe object hallucinations common in other baselines.

\noindent\textbf{Failure Cases.} 
Despite the significant improvements, affective reasoning in visual art remains challenging, occasionally leading to partial hallucinations. The second example (bottom of Figure~\ref{fig:app_qualitative_cases}) highlights a typical failure mode of our framework. The artwork is a somber sketch of a mother with her children. The baselines again suffer from severe hallucinations: AffectGPT imagines ``pain, mouth open wide'' and ``{with hands clasped together}'', while both Emotion-llama and Videmo bizarrely hallucinate that the woman is ``holding a cat''. 

ProFocus successfully predicts the correct emotion (``Sadness'') and correctly identifies the main subject (``woman is sitting on the chair''). However, to rationalize the predicted negative emotion, the language decoder hallucinates a specific action, claiming that the ``baby is crying''. In reality, the sadness in the artwork stems from a more abstract, implicit sense of exhaustion (e.g., ``a mother's work is never done''). This failure indicates that while hierarchical priors provide accurate structural anchors, the language decoder can sometimes over-rely on explicit negative visual actions (like crying) to justify complex, implicit emotional atmospheres.

\end{document}